\pdfoutput=1

\documentclass[11pt]{article}

\usepackage{ACL2023}

\usepackage{amsmath,amsfonts,bm}

\usepackage{xspace}
\usepackage[nameinlink]{cleveref}

\crefformat{section}{\S#2#1#3} 
\crefname{algorithm}{Alg.}{Algs.}
\crefformat{subsection}{\S#2#1#3}
\Crefname{equation}{Eq.}{Eqs.}
\Crefname{figure}{Fig.}{Figs.}

\newcommand{\ie}{{\em i.e.,}\xspace}
\newcommand{\cf}{{\em c.f.,}\xspace}
\newcommand{\eg}{{\em e.g.,}\xspace}

\newcommand{\wrt}{\emph{w.r.t.}\xspace}
\newcommand{\aka}{\emph{a.k.a.}\xspace}

\newcommand{\Ni}{({\em i})~}
\newcommand{\Nii}{({\em ii})~}
\newcommand{\Niii}{({\em iii})~}
\newcommand{\Niv}{({\em iv})~}

\def\eqref#1{equation~\ref{#1}}

\def\1{\bm{1}}

\def\rvg{{\mathbf{g}}}

\DeclareMathAlphabet{\mathsfit}{\encodingdefault}{\sfdefault}{m}{sl}
\SetMathAlphabet{\mathsfit}{bold}{\encodingdefault}{\sfdefault}{bx}{n}

\def\gN{{\mathcal{N}}}

\def\gY{{\mathcal{Y}}}

\DeclareMathOperator*{\argmax}{arg\,max}

\usepackage{times}
\usepackage{latexsym}

\usepackage[T1]{fontenc}

\usepackage[utf8]{inputenc}

\usepackage{microtype}

\usepackage{inconsolata}

\usepackage{algorithm,algpseudocode}
\usepackage{graphicx}
\usepackage{booktabs}       

\newcommand{\info}{\citep{wang2021infobert}}
\newcommand{\free}{\citep{zhu2020freelb}}
\newcommand{\safer}{\citep{safer20}}
\newcommand{\adam}{\citep{loshchilov2019decoupled}}

\usepackage{multirow}
\newtheorem{proof}{Proof}
\newtheorem{theorem}{Theorem}
\newtheorem{lemma}{Lemma}

\newtheorem{definition}{Definition}

\title{Randomized Smoothing with Masked Inference for \\ Adversarially Robust Text Classifications}

\author{Han Cheol Moon$\clubsuit$, Shafiq Joty$\clubsuit \lozenge$, Ruochen Zhao $\clubsuit$, Megh Thakkar$\dagger$, and Xu Chi$\spadesuit$\\
  $\clubsuit$Nanyang Technological University, Singapore \\ $\lozenge$Salesforce AI\\
  $\dagger$ Birla Institute of Technology and Science, Pilani\\ $\spadesuit$A*STAR, Singapore \\
\texttt{\{hancheol001@e., sjoty@, ruochen002@e.\}ntu.edu.sg} }

\begin{document}
\maketitle
\begin{abstract}
	Large-scale pre-trained language models have shown outstanding performance in a variety of NLP tasks. However, they are also known to be significantly brittle against specifically crafted adversarial examples, leading to increasing interest in probing the adversarial robustness of NLP systems. We introduce RSMI, a novel two-stage framework that combines randomized smoothing (RS) with masked inference (MI) to improve the adversarial robustness of NLP systems. RS transforms a classifier into a smoothed classifier to obtain robust representations, whereas MI forces a model to exploit the surrounding context of a masked token in an input sequence. RSMI improves adversarial robustness by \textbf{2 to 3 times} over existing state-of-the-art methods on benchmark datasets. We also perform in-depth qualitative analysis to validate the effectiveness of the different stages of RSMI and probe the impact of its components through extensive ablations. By empirically proving the stability of RSMI, we put it forward as a practical method to robustly train large-scale NLP models. Our code and datasets are available at \url{https://github.com/Han8931/rsmi_nlp}. 
\end{abstract}

\section{Introduction}

\begin{figure}[h]
	\centering
	\includegraphics[scale=0.91]{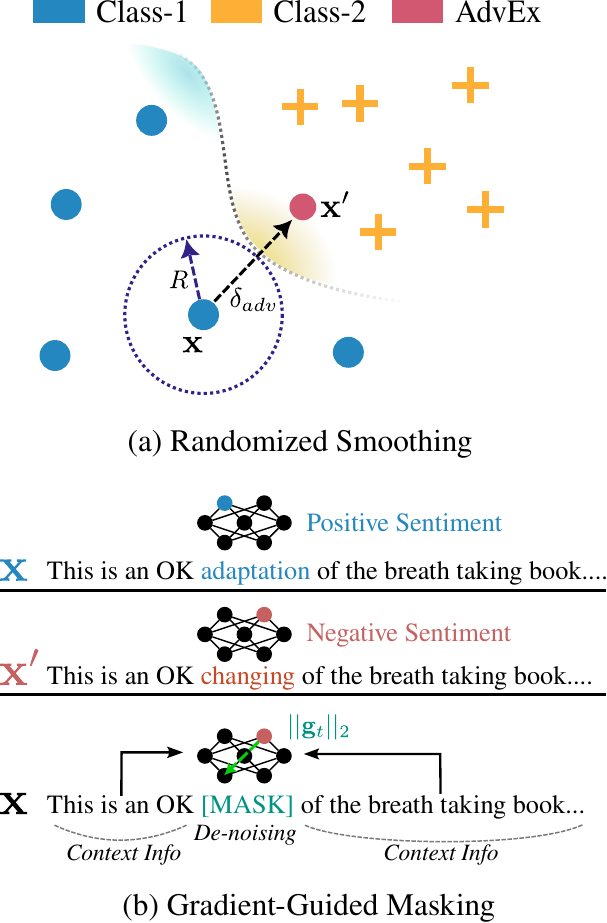}
	\caption{An overview of RSMI. (a) Randomized smoothing (RS) provides a \textit{certifiable robustness} within a ball with a radius $R$ around an input point $\mathbf{x}$ (\cf $R$ can be computed by \Cref{thrm:newradius_modified}) and (b) masked inference (MI) \textit{denoises} ``adversarially salient'' tokens via a gradient-based feature attribution analysis to make a decision on the input sample with the surrounding contexts of a masked token in the input.}
	\label{fig:rsmi_overview}
\end{figure}

In response to the threat of textual adversarial attacks \citep{Ebrahimi2018-hotflip,jin2019textfooler}, a variety of defense schemes have been proposed \citep{advNLPsurvey,safer20,Zhou21-DNE,Dong2021}. Defense schemes typically involve solving a min-max optimization problem, consisting of an \textit{inner maximization} that seeks the worst (adversarial) example and an \textit{outer minimization} that aims to minimize a system's loss over such examples \citep{madry2018towards}. The solution to the inner maximization problem is typically obtained through iterative optimization algorithms, such as stochastic gradient descent \citep{Ilyas2019,madry2018towards}. 

For texts, however, the gradients cannot be directly computed due to the discrete nature of the texts. Thus, the gradients are often computed with respect to word embeddings of an input sequence as done in \citet{miyato2016adversarial,zhu2020freelb,wang2021infobert}. The simplicity of the gradient-based adversarial training makes it attractive defense strategy, but it tends to show substantial variance in robustness enhancement (\cf \Cref{sec:train_stability}). Another prevailing approach in natural language processing (NLP) field is to substitute input words of their synonyms sampled from a pre-defined synonym set \citep{safer20,Zhou21-DNE,Dong2021}. The synonym-based defense (SDA) algorithms have emerged as a prominent defense approach since many textual attack algorithms perturb input texts at a word level \citep{pwws2019,alzantot-etal-2018-generating,jin2019textfooler}. However, \citet{li-etal-2021-searching} pointed out that they tend to {show significant brittleness} when they have no access to the perturbation sets of the potential attacks. In addition, assuming access to the potential perturbation sets is often unrealistic.

To address the issues, we propose RSMI, a novel two-stage framework leveraging randomized smoothing (RS) and masked inference (MI) (\cf \Cref{fig:rsmi_overview}). \textit{Randomized smoothing} is a generic class of methods that transform a classifier into a smoothed classifier via a randomized input perturbation process \citep{Cohen19_RS,Lecuyer19_DP}. It has come recently into the spotlight thanks to its simplicity and theoretical guarantee of \textit{certifiable robustness} within a ball around an input point \citep{Cohen19_RS,Salman19_PRS}, which is often considered desirable property of a defense scheme (\cf \Cref{fig:rsmi_overview} (a)). Moreover, its robustness enhancement is highly \textit{scalable} to modern large-scale deep learning setups \citep{Cohen19_RS,Lecuyer19_DP}. These properties render it a promising research direction. However, there exists a non-trivial challenge in introducing RS to NLP systems due to the discrete nature of texts. We sidestep the issue and adapt RS to NLP problems by injecting noise at the hidden layers of the deep neural model. We show that perturbed representations of pre-trained language models (PLMs) still guarantees the robustness in \Cref{sec:rsmi_main}.

The RS stage is followed by a gradient-guided masked inference (MI) to further reinforce the smoothing effect of RS. MI draws an inference on an input sequence via a {noise reduction process that masks adversarially ``salient'' tokens} in the input that are potentially perturbed by an attack algorithm (\cf \Cref{fig:rsmi_overview} (b)). The adversarially salient tokens are achieved via a \textit{gradient-based feature attribution analysis} rather than random selections as commonly done in pre-training language models \citep{bert} to effectively suppress adversarial perturbations. The effectiveness of our novel MI can be attributed to several aspects: \Ni It is a {natural regularization} for forcing the model to make a prediction based on the surrounding contexts of a masked token in the input \citep{Moon2021_Masker}. \Nii It works {without any prior assumption} about potential attacks, which renders it an \textit{attack-agnostic} defense approach and is more practical in that it requires no sub-module for synonym-substitution. \Niii It has a close {theoretical connection to the synonym-substitution-based approaches}, as MI can be regarded as a special case of the weighted ensemble over multiple transformed inputs as shown in \Cref{subsec:grad_mask}.

We evaluate the performance of RSMI through comprehensive experimental studies on large-scale PLMs with three benchmark datasets against widely adopted adversarial attacks. Our empirical studies demonstrate that RSMI obtains improvements of \textbf{$\mathbf{2}$ to $\mathbf{3}$ times} against strong adversarial attacks in terms of key robustness metrics over baseline algorithms despite its simplicity (\Cref{subsec:main_exp}). We also conduct theoretical analysis to demonstrate the effectiveness of our adapted RS (\Cref{subsec:noise_layer}) and MI (\Cref{subsec:grad_mask,sec:masked_inf}). We further analyze the scalability of RSMI, the influence of hyperparameters, its impact on the latent representation (\ie embedding space) of the system and its stochastic stability (\Cref{subsec:eot}). Our theoretical and empirical analyses validate the effectiveness of RSMI and propose it as a practical method for training adversarially robust NLP systems.

\section{Randomized Smoothing with Masked Inference (RSMI)}
\label{sec:rsmi_main}

We consider a standard text classification task with a probabilistic model $F_{\boldsymbol{\theta}}:\mathbb{R}^{d}\to \mathcal{P}(\mathcal{Y})$, where $\mathcal{P}(\mathcal{Y})$ is the set of possible probability distributions over class labels $\mathcal{Y}=\{1, \ldots, C\}$, and $\boldsymbol{\theta}\in \mathbb{R}^{p}$ denotes the parameters of the model $F_{\boldsymbol{\theta}}$ (or $F$ for simplicity). The model $F$ is trained to fit a data distribution $\mathcal{D}$ over pairs of an input sequence $s = (s_1, \ldots, s_T)$ of $T$ tokens and its corresponding class label ${y}\in \mathcal{Y}$. The distributed representation of $s$ (or word embedding) is represented as $x = [x_1,\ldots,x_T]$. We assume the model is trained with a loss function $\mathcal{L}$ such as cross-entropy. We denote the final prediction of the model as $\hat{y} = \argmax_i F(s)_i$ and the ground truth label as $y^*$.

\subsection{Randomized smoothing via noise layers} 
\label{subsec:noise_layer}
Given the model $F$, our method exploits a \emph{randomized smoothing} \citep{Lecuyer19_DP, Cohen19_RS} approach to obtain a smoothed version of it, denoted by $G:\mathbb{R}^{d}\to \mathcal{P}(\mathcal{Y})$, which is provably robust under isotropic Gaussian noise perturbation $\delta$ at an input query $u$ {(\eg\ an image)}.  This can be expressed as:
\begin{definition}
Given an original probabilistic neural network classifier $F$, the associated smoothed classifier $G$ at a query $u$  can be denoted as (\aka\ \textit{Weierstrass Transform} \citep{zayed1996}):
\begin{align}
\begin{split}
	G(u) &=  (F*\mathcal{N}(0,\sigma^2I))(u)\\
	&= \mathbb{E}_{\delta\sim\mathcal{N}(0,\sigma^2I)}[F(u+\delta)]\,.
	\label{eq:smooth_model}
 \end{split}
\end{align}
\end{definition}
The standard deviation of the Gaussian noise $\sigma$ is a hyperparameter that controls the robustness/accuracy trade-off of the resulting smoothed model $G$. The higher the noise level is, the more robust it will be, while the prediction accuracy may decrease. The asterisk $*$ denotes a convolution operation \citep{Oppenheim1996SignalsSystems} which, for any two functions $h$ and $\psi$, can be defined as: $h*\psi(x) = \int_{\mathbb{R}^d}h(t)\psi(x-t)dt$. {In practice, $G(u)$ can be estimated via Monte-Carlo sampling \citep{Cohen19_RS, Salman19_PRS}.} 

\citet{Cohen19_RS} showed that the smoothed model $G$ is robust around a query point $u$ within a $L_2$ radius $R$, which is given by:
\begin{equation}
	R = \frac{\sigma}{2}(\Phi^{-1} (p_a) - \Phi^{-1} (p_b))\,,
	\label{eq:robust_radius}
\end{equation}
where $\Phi^{-1}$ is the inverse of the standard Gaussian CDF, $p_a$ and $p_b$ are the probabilities of the two most likely classes $a$ and $b$, denoted as: $a = \argmax_{y \in \gY} G(x)_y \text{ and }~ b = \argmax_{y \in \gY\backslash{a}} G(x)_y$.

As per \Cref{eq:smooth_model}, a simple approach to obtain $G$ is to perturb the input $u$ by the noise $\delta$ and train with it. However, for a textual input, the token sequence cannot be directly perturbed by $\delta$ due to the its discrete nature. To deviate from the issue, we inject noise at the hidden layers of the model to achieve stronger smoothness as done in \citep{Liu18_RSE}. For a given layer $f_l$, a noise layer $f^\delta_{l}$ draws a noise $\delta \sim \mathcal{N}(0,\sigma^2I)$ and adds it to the output of $f_l$ in {every forward pass of the model}. The stronger smoothness resulting from the multi-layer noise is provably guaranteed by the following theorem:

\begin{theorem}
\label{thrm:newradius_modified}
Let $F: \mathbb{R}^d \rightarrow \mathcal{P}(\mathcal{Y})$ be any soft classifier which can be decomposed as $F = f_1 \circ f_2 \circ \dots \circ f_L$ and $G = g_1 \circ g_2 \circ \dots \circ g_L$ be its associated smoothed classifier, where $g_l(x) = (f_l * N(0, \sigma_l^2 I ))(x)$ with $1\leq l \leq L$ and $\sigma_l >0$. Let $a = \argmax_{y \in \mathcal{Y}} G(x)_y$ and $b = \argmax_{y \in \mathcal{Y}\backslash{a}} G(x)_y$ be two most likely classes for $x$ according to $G$. Then, we have that $\argmax_{y \in \mathcal{Y}} G(x')_y = a$ for $x'$ satisfying
\begin{small}
	$$\left\lVert x' - x \right \rVert_2 \leq \frac{1}{2\sigma_1}\prod_{l=2}^L (1 + \sigma_l^2)(\Phi^{-1} (p_a) - \Phi^{-1} (p_b))\,.$$
\end{small}
\end{theorem}
We provide a proof of the theorem with \textit{Lipschitz continuity} in \Cref{sec:proof_smooth}.

\subsection{Gradient-guided masked inference}
\label{subsec:grad_mask}

For an input sequence $s$, our method attempts to \textit{denoise} its adversarially perturbed counterpart $s'$ by attributing saliency of input tokens through a simple gradient-based attribution analysis. Due to the discrete nature of tokens, we compute the gradients of the loss function $\mathcal{L}$ with respect to the word embeddings $x_t$. The loss is computed with respect to the labels $y$, which is set to be the ground-truth labels $y^*$ during training and model predictions $\hat{y}$ during inference. Formally, the gradients $\mathbf{g}_t$ for a token $s_t \in s$ (correspondingly $x_t \in x$) can be computed as follows: 
\begin{align}
\begin{split}
	\mathbf{g}_t &= \nabla_{x_t}\mathcal{L}(G(x), y)\\ 
	&\approx - \nabla_{x_t} \bigg(\log \bigg(\frac{1}{\nu} \sum_{i=1}^\nu G(x+\delta_i)\bigg)\bigg).
	\label{eq:smooth_grad}
 \end{split}
\end{align}
\Cref{eq:smooth_grad} exploits a Monte-Carlo approximation to estimate the gradient $\rvg_t$ as done in \citep{Salman19_PRS}. Subsequently, the amount of stimulus of the input tokens toward the model prediction is measured by computing the $L_2$-norm of $\mathbf{g}_t$, \ie $||\mathbf{g}_t||_2$. The stimulus is regarded as the \emph{saliency score} of the tokens and they are sorted in descending order of the magnitude \citep{li2016visualizing,gradmask}. Then, we sample $M$ tokens from the top-$N$ tokens in $s$, and mask them to generate a masked input sequence $m = [s_1,\ldots, m_t, \ldots, s_T]$, where $t$ is the position of a salient token and $m_t$ is the mask token, $[\textrm{MASK}]$. During training, we mask the top-$M$ positions (\ie $N=M$), while the mask token selection procedure is switched to a sampling-based approach during inference as detailed later in \Cref{subsec:two_step_inf}. Finally, the gradients $\rvg_t$ computed for generating the masked sequence is repurposed for perturbing the word embeddings $x_t$ (\ie $\delta = \rvg_t$) to obtain robust  embeddings as shown in \citep{zhu2020freelb, wang2021infobert, shen2018deep}. 

Our gradient-guided masked inference offers several advantages. First, it yields a natural regularization for forcing the model to exploit surrounding contexts of a masked token in the input \citep{Moon2021_Masker}. Second, the masking process can provide a better smoothing effect by masking `salient' tokens that are probably adversarial to the model's decision. In such cases, it works as a denoising process for reducing the strength of an attacks. In \Cref{sec:masked_inf}, we conduct theoretical analysis about the denoising effect of the gradient-guided masked inference in terms of Lipschitz continuity of a soft classifier.

\begin{algorithm*}[t!]
	\begin{algorithmic}[1]
		\State Initialize. $s$, $M$, $N$, $\sigma$, $\nu$, $k_0$, $k_1$, $\alpha$, step size $\eta$, and a gradient scale parameter $\beta$.
		\State Compute $L:=\big[||\mathbf{g}_1||_2,\cdots, ||\mathbf{g}_T||_2\big]$  via \Cref{eq:smooth_grad}. \Comment{Gradients \wrt\ word embeddings}
		\State Sort $L$ in descending order and keep top-$N$ items 
		\State Get a masked sequence $m$ by masking top-$M$ tokens based on $L$.
		\If {Training}
		\State $x:=x+\beta(\mathbf{g}_1,\cdots,\mathbf{g}_T )$ \Comment{Noise to word embeddings}
		\State $\boldsymbol{\theta}:=\boldsymbol{\theta}-\eta\nabla_{\boldsymbol{\theta}}\mathcal{L}(G(x), y^*)$ 
		\ElsIf{Prediction}
		\State $\boldsymbol{\phi}(m)_0 = \sum_{i=1}^{k_0}[\mathbb{I}(\hat{y}^{(i)}(m)=y_1),\cdots,\mathbb{I}(\hat{y}^{(i)}(m)=y_c) ]$   \Comment{First vote}
		\State $n_a = \max \boldsymbol{\phi}(m)_0$ 
		\State $p$-value = \textsc{BinomTest}($n_a,k_0,0.5$,one-tail)
		\If{$p$-value $>\alpha$}
		\State Return $\argmax_{y\in \mathcal{Y}} \boldsymbol{\phi}(m)_0$ 
		\Else
		\State $[m^{(1)},\cdots, m^{(k_1)}]\sim$ \textsc{RandGradMask($k_1,L$)}
		\State $\boldsymbol{\phi}(m)_1 = \sum_{i=1}^{k_1}[\mathbb{I}(\hat{y}(m^{(i)})=y_1),\cdots,\mathbb{I}(\hat{y}(m^{(i)})=y_c) ]$ \Comment{Second vote}
		\State Return $\argmax_{y\in \mathcal{Y}} \boldsymbol{\phi}(m)_1$
		\EndIf
	\EndIf
	\end{algorithmic}
	\caption{Training and prediction procedure of RSMI.}
	\label{alg:rsmi}
\end{algorithm*}

\paragraph{Connection to synonym-based defense methods}
Another interesting interpretation is that the masked inference has a close connection to the synonym-based defense methods \citep{Wang2021-DP,safer20,Wang20-RSE,Zhou21-DNE}. Assuming only position in $s$ is masked and treating the mask as a latent variable $\tilde{s}_{t}$ that could take any token from the vocabulary $V$, we can express the masked inference as: 
\begin{align}
	p(y|m) &= \sum_{\tilde{s}_{t} \in V} p(y, \tilde{s}_{t}|m) \\
	&=  \sum_{\tilde{s}_{t} \in V} p(y|m, \tilde{s}_{t})p(\tilde{s}_{t}|m)\\ 
	&\approx \sum_{\tilde{s}_{t} \in V_t} p(y|m, \tilde{s}_{t})p(\tilde{s}_{t}|m)\,,
	\label{eq:mask_lm}
\end{align}
where $|V_t|\ll|V|$ is the number of words to be at position $t$ with a high probability mass. As shown in the equation, the masked inference can be factorized into a classification objective and a masked language modeling objective, which can be further approximated into a weighted ensemble inference with $|V_t|$ substitutions of $s$ with the highly probable tokens (\eg synonyms) corresponding to the original word $s_t$. If we assume $p(\tilde{s}_{t}|m)$ to be a probability of sampling uniformly from a synonym set such as the one from the WordNet \citep{wordnet}, then the masked inference is reduced to a synonym-substitution based defense approach with no necessity of an explicit synonym set.

\subsection{Two-step Monte-Carlo sampling for efficient inference}
\label{subsec:two_step_inf}

The prediction procedure of RSMI involves averaging predictions of $k$ Monte-Carlo samples of $G(m)$ to deal with the variations in the noise layers. A large number of $k$ is typically required for a robust prediction but the computational cost increases proportionally as $k$ gets larger. To alleviate the computational cost, we propose a two-step sampling-based inference (\Cref{alg:rsmi}). 

In the first step, we make $k_0$ predictions by estimating $G(m)$ for $k_0$ times ($k_0$ forward passes). We then make an initial guess about the label of the masked sample $m$ by taking a majority vote of the predictions. Following \citet{Cohen19_RS}, this initial guess is then tested by a one-tailed binomial test with a significance level of $\alpha$. If the guess passes the test, we return the most probable class based on the vote result. If it fails, then we attempt to make a second guess with a set of $k_1$ masked input sequences $\mathcal{M}=[m^{(1)},\cdots, m^{(k_1)}]$, where $k_0\ll k_1$. Note that the masked input $m$ used in the first step is generated by masking the top-$M$ tokens from the top-$N$ candidates as we do during training. However, in the second step, we randomly sample $M$ masking positions from the $N$ candidates to create each {masked sequence} $m^{(i)}$ of $\mathcal{M}$ in order to maximize variations in the predictions; this step is denoted as \textsc{RandGradMask} in \Cref{alg:rsmi}.

Our two-step sampling based inference is based on an assumption that textual adversarial examples are liable to fail to achieve consensus from the RSMI's predictions compared to clean samples. In other words, it is considerably harder for adversarial examples to estimate the optimal perturbation direction towards the decision boundary of a stochastic network \citep{Sina2022, athalye2018obfuscated, Cohen19_RS}. We conduct experiments to show the effectiveness of our two-step sampling based inference in \Cref{subsec:eot}.

\section{Experiment Setup}
\label{sec:exp_setup}
\begin{table*}[t]
	\setlength{\tabcolsep}{2.0pt}
	\centering
 	\scalebox{0.60}{
	\begin{tabular}{lllc|cccc|cccc|ccccccc}
	\toprule
\bf Dataset & \bf PLM &\bf Model & \bf SAcc ($\uparrow$) & \multicolumn{4}{c}{\bf RAcc  ($\uparrow$)} & \multicolumn{4}{c}{\bf ASR ($\downarrow$)}&  \multicolumn{4}{c}{\bf AvgQ ($\uparrow$)}  \\
&  & & &  TF&PWWS&BAE & Avg. & TF&PWWS&BAE & Avg. & TF&PWWS&BAE  & Avg.\\ 
	\midrule
	\multirow{14}{*}{IMDb}
&\multirow{7}{*}{BERT-base}
& + Fine-Tuned         &90.60& 5.90  & 0.60  & 27.30 & 11.27 & 93.49 & 99.34 & 69.87 & 87.57 & 440  & 1227 & 377   & 681  \\
&& + FreeLB \free      &92.90& 10.50 & 14.22 & 45.30 & 23.34 & 88.70 & 84.71 & 51.24 & 74.88 & 909  & 1400 & 442   & 917  \\
&& + InfoBERT \info    &\bf 92.90& 26.40 & 26.80 & 50.00 & 34.40 & 71.58 & 71.15 & 46.18 & 62.97 & 1079 & 1477 & 458   & 1005 \\
&& + SAFER \safer      &91.80& 23.80 & 30.90 & 38.20 & 30.97 & 74.08 & 66.34 & 58.39 & 66.27 & 1090 & 1504 & 618   & 1071 \\
&& + AdvAug            &92.60& 32.00 & 36.60& 52.10 & 40.23 & 65.44 & 60.48 & 43.74 & 56.55 & 1291	& 1569	& 478 & 1113\\
&& + RSMI-NoMask (Our) &91.00& 34.90 & 57.00 & 58.40 & 50.10 & 61.65 & 37.36 & 35.83 & 44.95 & 1395 & 1733 & 817   & 1315 \\
&& + RSMI (Our)        &92.20& \bf 56.40 & \bf 58.70 & \bf 80.20 & \bf 65.10 & \bf 38.83 & \bf 36.34 & \bf 13.02 & \bf 29.40 & \bf 1651 & \bf 1764 & \bf 1287 & \bf 1567 \\

	\cmidrule(l){2-16}
&\multirow{7}{*}{RoBERTa-base}
& + Fine-Tuned         &93.10& 0.50  & 1.10  & 22.60 & 8.07  & 99.46 & 98.82 & 75.73 & 91.34 & 588  & 1248 & 398   & 745  \\
&& + FreeLB \free      &93.20& 17.30 & 21.20 & 49.50 & 29.33 & 81.44 & 77.25 & 46.89 & 68.53 & 999  & 1433 & 461   & 964  \\
&& + InfoBERT \info    &94.00& 7.60  & 13.20 & 36.10 & 18.97 & 91.92 & 85.96 & 61.60 & 79.83 & 855  & 1388 & 418   & 887  \\
&& + SAFER \safer      &93.20& 31.80 & 39.20 & 45.40 & 38.80 & 65.88 & 57.94 & 51.29 & 58.37 & 1276 & 1575 & 678   & 1176 \\
&& + AdvAug            &\bf 94.40 & 28.90 & 31.60& 51.40 & 37.30 & 69.39 & 66.53 & 45.55 & 60.49 & 1220	& 1567	& 479 & 1089\\
&& + RSMI-NoMask (Our) &93.30& 47.00 & 54.00 & 52.10 & 51.03 & 49.63 & 42.12 & 44.16 & 45.30 & 1455 & 1684 & 764   & 1301 \\
&& + RSMI (Our)        &93.00& \bf 73.40 &\bf 76.20 &\bf 83.00 &\bf 77.53 &\bf 21.08 &\bf 18.07 &\bf 10.75 &\bf 16.63 &\bf 1917 &\bf 1863 &\bf 1314 &\bf  1698 \\

	\midrule
	\multirow{14}{*}{AGNews}  
&\multirow{7}{*}{BERT-base}
 & + Fine-Tuned         &93.90& 16.80 & 34.00 & 81.00 & 43.93 & 82.11 & 63.79 & 13.74 & 53.21 & 330  & 352  & 124   & 269  \\
&& + FreeLB \free      &\bf95.00& 24.40 & 48.20 & 84.10 & 52.23 & 74.32 & 49.26 & 11.47 & 45.02 & 383  & 367  & 131   & 294  \\
&& + InfoBERT \info    &94.81& 19.90 & 40.90 & 84.90 & 48.57 & 79.01 & 56.86 & 10.45 & 48.77 & 371  & 365  & 126   & 287  \\
&& + SAFER \safer      &93.70& 46.30 & 64.00 & 80.00 & 63.43 & 50.59 & 31.70 & 14.62 & 32.30 & 447  & 379  & 170   & 332  \\
&& + AdvAug            & 93.90 & 54.90 & 66.00& 80.10 & 67.00 & 41.53 & 29.71 & 14.70 & 28.65 & 465	& 386	& 129 & 327 \\
&& + RSMI-NoMask (Our) &92.60& 60.40 & 75.30 & 77.90 & 71.20 & 34.77 & 18.68 & 15.88 & 23.11 & 497  & 395  & 203   & 365  \\
&& + RSMI (Our)        &92.70&\bf 63.20 &\bf 76.10 &\bf 86.10 &\bf 75.13 &\bf 31.82 &\bf 17.91 &\bf 7.12  &\bf 18.95 &\bf 503  &\bf 397  &\bf 573   &\bf 491  \\
	\cmidrule(l){2-16}
&\multirow{7}{*}{RoBERTa-base}
 & + Fine-Tuned         &93.91& 23.90 & 49.30 & 80.00 & 51.07 & 74.55 & 47.50 & 14.80 & 45.62 & 353  & 367  & 130   & 283  \\
&& + FreeLB \free      &\bf95.11& 23.90 & 48.20 & 83.00 & 51.70 & 74.87 & 49.32 & 12.73 & 45.64 & 393  & 374  & 127   & 298  \\
&& + InfoBERT \info    &94.00& 30.20 & 52.30 & 79.80 & 54.10 & 67.87 & 44.36 & 15.11 & 42.45 & 396  & 374  & 134   & 301  \\
&& + SAFER \safer      &93.60& 49.30 & 68.90 & 81.60 & 66.60 & 47.33 & 26.39 & 12.82 & 28.85 & 452  & 386  & 172   & 337  \\
&& + AdvAug            & 94.00 & 61.00 & 70.90& 81.30 & 71.07 & 35.11 & 24.57 & 13.51 & 24.40 & 486	& 388	& 133 & 336 \\
&& + RSMI-NoMask (Our) &94.10& 66.40 & 79.00 & 82.80 & 76.07 & 29.44 & 16.05 & 12.01 & 19.17 & 504  & 396  & 213   & 371  \\
&& + RSMI (Our)        &94.30&\bf 74.10 &\bf 81.90 &\bf 88.60 &\bf 81.53 &\bf 21.42 &\bf 13.15 &\bf 6.04  &\bf 13.54 &\bf 530  &\bf 401  &\bf 576   &\bf 502  \\
	\bottomrule
	\end{tabular}
 	}
	\caption{Performance comparison of adversarial robustness of RSMI with the baselines for classification tasks. RSMI-NoMask excludes masking during inference time. Avg. stands for an average of evaluation results. }
	\label{table:robust_eval}
\end{table*}

\paragraph{Datasets}
We evaluate RSMI on two conventional NLP tasks: text CLaSsification (CLS) and Natural Language Inference (NLI). We adopt \textsc{IMDb} \citep{imdb} and \textsc{AG's News} \citep{zhang2016characterlevel} datasets for the classification task. For NLI, we compare the defense algorithms on the Question-answering NLI (QNLI) dataset, which is a part of the GLUE benchmark \citep{wang-etal-2018-glue}. We build development sets for \textsc{IMDb}, \textsc{AG}, and QNLI by randomly drawing 10\% samples from each training set via a stratified sampling strategy.

\paragraph{Evaluation metrics} \label{subsec:metrics}
The performance of defense algorithms is evaluated in terms of four different metrics as proposed in \citep{li-etal-2021-searching}: \Ni Standard accuracy (SAcc) is the model's accuracy on clean samples. \Nii Robust accuracy (RAcc) measures the model's robustness against adversarial attacks. \Niii Attack success rate (ASR) is the ratio of the inputs that successfully fool the victim models. \Niv Finally, the average number of queries (AvgQ) needed to generate the adversarial examples. 

\paragraph{Baselines}
We select FreeLB \citep{zhu2020freelb}\footnote{FreeLB++ \citep{li-etal-2021-searching} is excluded since it has a reproducibility issue as reported in \href{https://github.com/RockyLzy/TextDefender}{github}.}, InfoBERT \citep{wang2021infobert}, and an adversarial example augmentation (AdvAug) \citep{li-etal-2021-searching} as baselines since they are representative work for gradient-based training, information-theoretical constraint, data augmentation approaches, respectively. We also choose SAFER \citep{safer20} since it is a certifiable defense method based on a synonym-substitution-based approach. We then apply the baselines over BERT-base \citep{bert} and RoBERTa-base \citep{roberta} models. We also conduct experiments with RoBERTa-Large, BERT-Large, and {T5-Large} \citep{t5_model} models to observe scalability of RSMI. Note that our experiment setup is significantly larger compared to previous works, including \citet{li-etal-2021-searching,safer20}. The baseline algorithms are tuned according to their default configurations presented in the respective papers and run them three times with a different random initialization to obtain the best performance of the baselines. For AdvAug, we augment a training dataset of each task by adversarial examples sampled from 10k data points of training datasets. Further details are provided in \Cref{sec:parameter_setting}.

\paragraph{Textual adversarial attacks}
We generate adversarial examples via TextFooler (TF) \citep{jin2019textfooler}, Probability Weighted Word Saliency (PWWS) \citep{pwws2019} and BERT-based Adversarial Examples (BAE) \citep{garg-ramakrishnan-2020-bae}. These attack algorithms are widely adopted in a variety of defense works as adversarial robustness benchmarking algorithms \citep{yoo-qi-2021-towards-improving,Dong2021} since they tend to generate adversarial examples with better retention of semantics and show high attack effectiveness compared to syntactic paraphrasing attacks \citep{iyyer-etal-2018-adversarial}. Moreover, the above attack algorithms have their own distinct attack process. For instance, TF and PWWS build synonym sets by counter-fitting word embeddings \citep{mrksic-etal-2016-counter} and WordNet \citep{wordnet}, respectively. BAE leverages BERT for building a synonym set of a target token. Note that we exclude some adversarial attack algorithms, such as BERT-Attack \citep{li-etal-2020-bert-attack} due to their expensive computation costs \footnote{BERT-Attack takes around 2k times more than TextFooler algorithm to generate a single adversarial example of a AGNews sample under our experiment setup. This issue is also reported in \href{https://textattack.readthedocs.io/en/latest/3recipes/attack_recipes.html\#bert-attack}{TextAttack} \citep{morris2020textattack}.}. 

We randomly draw 1,000 samples from each test set following \citet{Dong2021,li-etal-2021-searching,safer20} for a fair comparison and perturb them via an attack to generate the corresponding adversarial examples for all experiments unless stated otherwise. The sample size is also partially due to the slow attack speed of textual adversarial attack algorithms and the strong robustness of the proposed model, which requires the attack algorithms to query a huge amount of times compared to the baselines (\cf \Cref{table:robust_eval}). We implement all attacks through the publicly available TextAttack library \citep{morris2020textattack} and use their default configurations without any explicit attack constraints. 

For robustness evaluation of RSMI against the attacks, we modify the second step of the two-step inference to make a final decision by averaging logit scores of $k_1$ Monte-Carlo samples instead of the majority voting approach in \Cref{alg:rsmi}. We do this to prevent obfuscating the perturbation processes of TF and PWWS that are devised to identify target tokens via the change of the model's confidence, which can give a false impression about the robustness of RSMI. Nonetheless, we investigate the effectiveness of majority voting based inference as a practical defense method in \Cref{sec:majority}. Further details about the attack algorithms and parameter settings of the algorithms are provided in \Cref{sec:parameter_setting}.

\section{Results and Analysis}
\subsection{Adversarial robustness comparison}
\label{subsec:main_exp}

\begin{table*}[t]
	\setlength{\tabcolsep}{4.5pt}
	\centering
 	\scalebox{0.90}{
	\begin{tabular}{llrrrr}
	\toprule
	\bf Dataset &\bf  Model &\bf  SAcc ($\uparrow$) &\bf  RAcc  ($\uparrow$) &\bf  ASR ($\downarrow$) &\bf  AvgQ ($\uparrow$) \\
	\midrule
	\multirow{3}{*}{IMDb}
	& BERT-Large + RSMI    & 93.16(+0.66) & 79.30(+49.80) & 14.88(-53.23) & 1980(+850) \\ 
	& RoBERTa-Large + RSMI & 95.06(+0.76) & 87.40(+66.20) & 8.06 (-69.46) & 2092(+1058) \\ 
	& T5-Large + RSMI & 94.41(+0.38) & 62.87(+35.24) & 33.40(-37.21) & 1684(+598)\\
	\midrule                                                               
	\multirow{3}{*}{AGNews}                                                
	& BERT-Large + RSMI    & 94.60(-0.70) & 85.70(+65.10) & 9.41 (-68.97) & 568 (+210)\\ 
	& RoBERTa-Large + RSMI & 94.60(+0.54) & 88.10(+46.60) & 6.87 (-49.01) & 577 (+144)\\
	& T5-Large + RSMI & 94.90(-0.10) & 75.10(+13.56) & 20.86(-14.37) & 516(+89) \\
	\bottomrule
	\end{tabular}
 	}
	\caption{Performance of adversarial robustness of RSMI on large-scale PLMs. The round brackets next to each number denote the change of score compared to its fine-tuned model.}
	\label{table:mlm_large}
\end{table*}

We compare the performance of RSMI with the baselines in \Cref{table:robust_eval}. Overall, we observe that \textbf{RSMI} outperforms all the baselines by quite a large margin across the majority of the metrics such as RAcc, ASR and AvgQ. In particular, it achieves about \textbf{2 to 3 times} improvements against strong attack algorithms (\eg TextFooler and PWWS) in terms of ASR and RAcc, which are key metrics for evaluating the robustness of defense algorithms. RSMI also significantly outperforms the baselines in QNLI task by $16\%\sim 26\%$ (\cf \Cref{sec:nli_task}). In addition, we observe that RSMI tends to show better training stability in RAcc compared to the baselines (\cf \Cref{sec:train_stability}). For instance, the RAcc of FreeLB shows a standard deviation of 8.57\%, but RSMI shows 2.10\% for the IMDb task. Also, InfoBERT tuned for AGNews shows a standard deviation of 13.19\% in RAcc while RSMI shows 0.84\%. We also emphasize that RSMI outperforms other existing methods, such as TAVAT \citep{tavat}, MixADA \citep{AdvMixUp}, A2T \citep{yoo-qi-2021-towards-improving}, and ASCC \citep{Dong2021} which we do not report in \Cref{table:robust_eval} as they show lower performance than our baselines, \cf \citet{li-etal-2021-searching}.\footnote{\citet{li-etal-2021-searching} put constraints to make the attack algorithms weaker which we did not do in our work.} Another interesting observation is that a simple AdvAug approach outperforms sophisticated methods, including InfoBERT, FreeLB, and SAFER in most experiment setups without hurting SAcc. This runs contrary to the claims in \citet{li-etal-2021-searching, AdvMixUp}. 

The strong performance of RSMI can be attributed to four factors: \Ni A provable robustness guarantee by the randomized smoothing approach helps attain higher robustness (\cf \Cref{thrm:newradius_modified}). To further support this claim, we evaluate the robustness of RSMI without the proposed masking process (\ie MI) during inference and the results are reported as RSMI-NoMask in \Cref{table:robust_eval}. As we can see, RSMI-NoMask  outperforms the baselines in most experiment scenarios. \Nii The randomized smoothing denoises adversarial perturbations in the latent space of systems. Our experiments in \Cref{subsec:rsmi_space} bolster this claim by showing the significant noise reduction in hidden representations of RSMI. \Niii The MI leads to a reduction in the noise of the adversarial perturbations. This claim can be strongly bolstered again by comparing the results of RSMI with and without the masking strategy during inference (\cf RSMI-NoMask and RSMI in \Cref{table:robust_eval}). \Niv The two-step sampling-based inference makes it harder for the attack algorithms to estimate the optimal {perturbation direction to fool the model for an input sample}. An ablation study in \Cref{subsec:eot} clearly supports this claim since the two-step sampling significantly improves the RAcc of RSMI models.

\subsection{Large scale parameterization and adversarial robustness}
We investigate the scalability of RSMI by applying RSMI over large-scale PLMs, such as RoBERTa-Large and BERT-Large, both of which have 340 million parameters. We also conduct experiments with T5-Large model of 770 million parameters. Then, we evaluate their robustness via TextFooler attack algorithm since it tends to show high ASR (\cf \Cref{table:robust_eval}). \Cref{table:mlm_large} summarizes the experiment results. From \Cref{table:mlm_large}, we can clearly observe that RSMI significantly improves the robustness of the large PLMs, which indicates the high scalability of RSMI. Especially, RoBERTa-Large with RSMI enhances the RAcc of the fine-tuned RoBERTa-Large by 66.20\% for the IMDb task.

\subsection{Analysis of latent representations of RSMI}
\label{subsec:rsmi_space}

\begin{figure}
     \centering
     \includegraphics[scale=0.58]{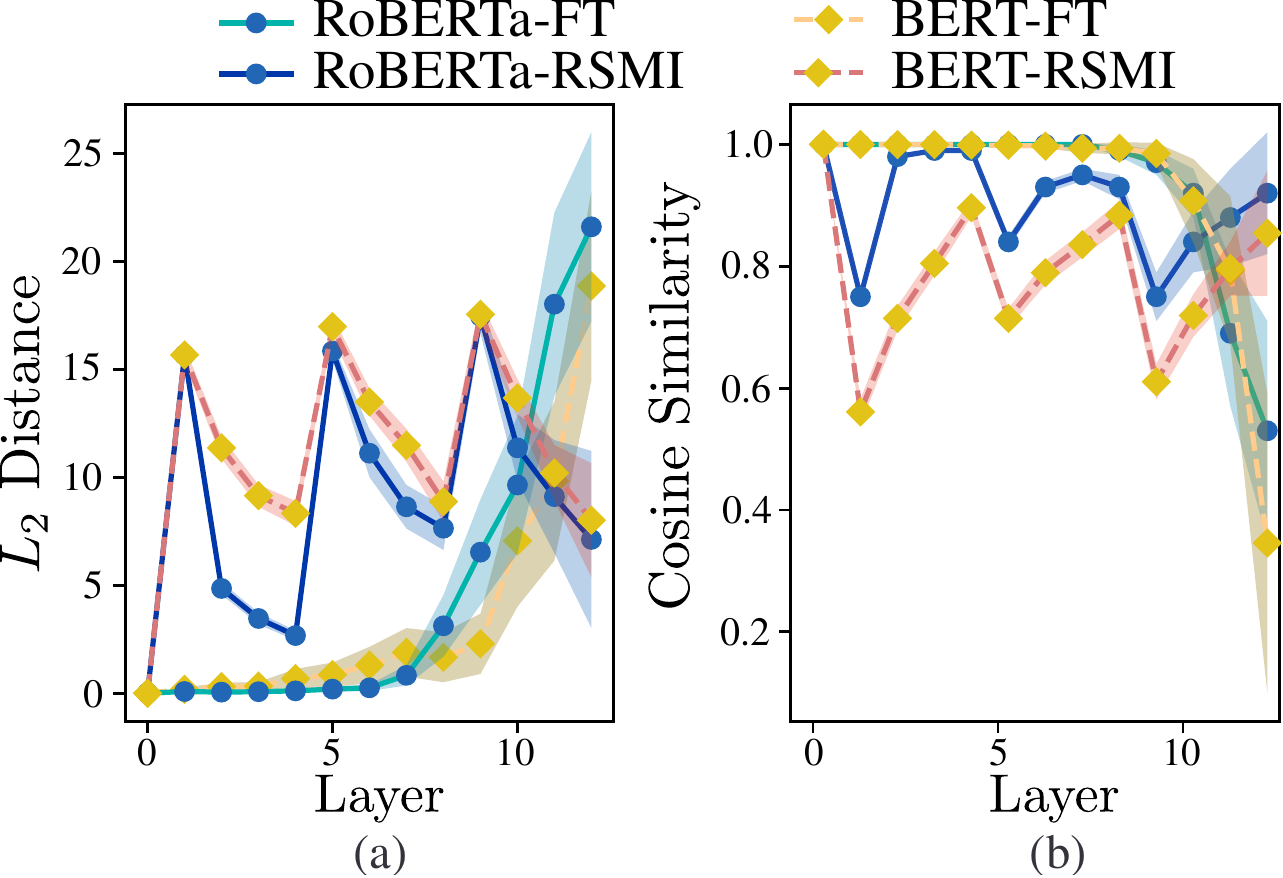}
	 \caption{Analysis of hidden representations of RSMI. We compare the $L_2$ distance and cosine similarity between a hidden representation of clean sample and that of its corresponding adversarial example. }
     \label{fig:hidden_repr}
 \end{figure}

 We investigate the latent representation of clean sample $h_l(s)$ and that of its corresponding adversarial example $h_l(s')$ for each layer $l$. We examine the fine-tuned base models and the base models with RSMI. For each model, we compare the $L_2$ distance and cosine similarity between $h_l(s)$ and $h_l(s')$ for each layer $l$. \Cref{fig:hidden_repr} shows that the $L_2$ distance and cosine similarity of the fine-tuned RoBERTa and BERT models stay quite constant until 8-th layer. However, for subsequent layers, $L_2$ distance curves rapidly increase and cosine similarity curves suddenly fall. At the final layer ($l=12$), we can observe the largest changes. Thus, the latent representation of the clean sample and its corresponding adversarial example become distinct. In contrast, RSMI tends to show significant decreases of $L_2$ distance at $l_1, l_5,$ and $l_9$ thanks to the Gaussian noise perturbation processes for these layers. This indicates that RSMI effectively reduces the adversarial noise in the latent representations of adversarial examples.

\subsection{Effectiveness of gradient-guided masking}
\label{sec:grad_mask}
\begin{table}[t]
	\centering
	\scalebox{0.85}{
	\begin{tabular}{cccccc}
    \toprule
	 & \bf Model &  \multicolumn{4}{c}{\bf ASR($\downarrow$)} \\ 
	 &       & $k=1$ & $k=5$ & $k=10$ & $k=50$ \\ 
	\midrule\multirow{2}{*}{$M=1$}
	 & RM  & 86.65 & 97.95 & 99.68 & 100\\ 
	 & GM  & \multicolumn{4}{c}{96.55} \\ 
	 \midrule\multirow{2}{*}{$M=2$}
	 &RM   & 76.57 & 90.84 & 93.11 & 96.27\\ 
	 &GM   & \multicolumn{4}{c}{83.12} \\ 
	 \midrule\multirow{2}{*}{$M=3$}
	 &RM   & 70.05 & 86.34 & 90.08 & 92.76 \\
	 &GM   & \multicolumn{4}{c}{90.81} \\ 
	 \midrule\multirow{2}{*}{$M=4$}
	 &RM   & 65.34  & 78.86  & 79.74 & 82.64 \\
	 &GM   & \multicolumn{4}{c}{81.34} \\ 
	 \midrule\multirow{2}{*}{$M=5$}
	 &RM   & 68.35 & 86.31  & 90.70 & 96.26\\
	 &GM   & \multicolumn{4}{c}{80.72} \\ 
	\bottomrule
	\end{tabular}
	}
	\caption{ ASR of random masking (RM) and gradient-guided masking (GM) for combinations of $M$ masked tokens and $k$ masked sequences.}
	\label{table:random_mask}
\end{table}

We probe the effectiveness of the gradient-guided masking strategy by ablating the noise layers and the two-step sampling of RSMI. The resulting model, namely the gradient-guided masking (GM) is compared to a model trained on randomly masked inputs, namely the random masking model (RM). Note that GM predicts and masks inputs in a deterministic way due to the absence of noise layers. \Cref{table:random_mask} summarizes our study of ASR changes of GM and RM over different number of mask tokens $M$ in an input sequence as well as $k$ randomly masked sequences drawn for estimating an expectation of RM prediction. RM tends to achieve its best performance at $k=1$, but shows vulnerability as $k$ increases, which means its robustness is largely from attack obfuscation rather than improving the model's robustness \citep{athalye2018obfuscated}. On the other hand, ASR of GM tends to decrease as we increase $M$. This {validates the effectiveness of gradient-guided masking for denoising adversarial perturbations} injected by attack algorithms.

\subsection{Effectiveness of two-step sampling}
\label{subsec:eot}
\begin{figure}[t]
    \centering
    \includegraphics[scale=0.68]{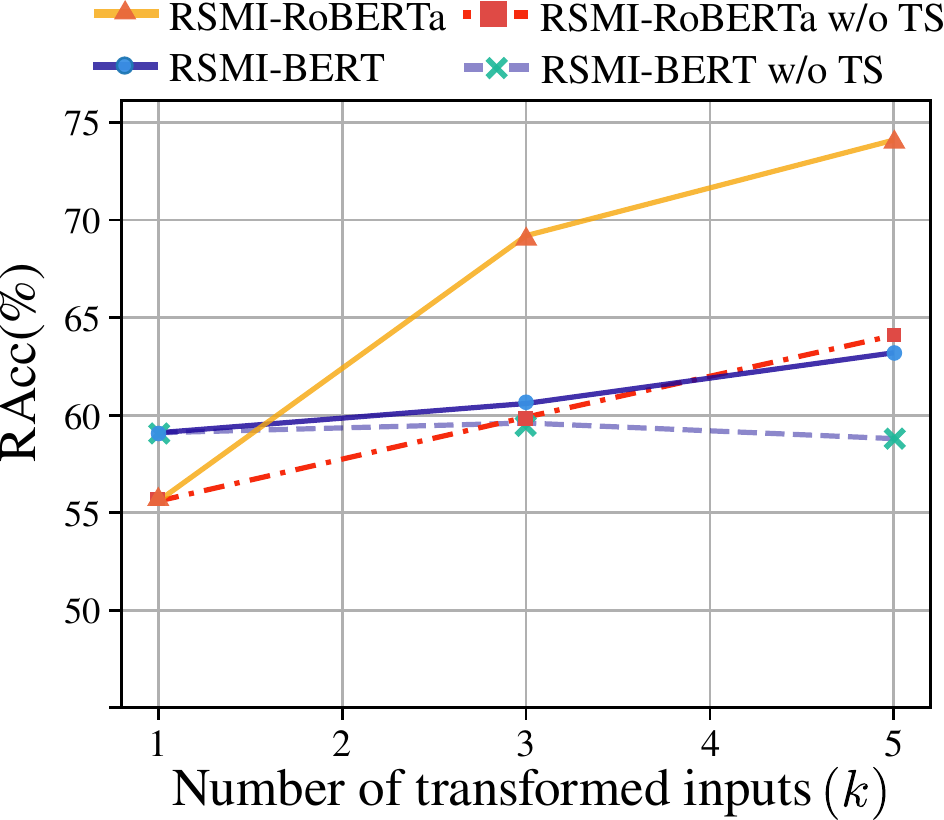}
	\caption{RAcc curves of RSMI with two-step (TS) sampling and without TS over input transformations.}
    \label{fig:ensemble}
\end{figure}

We study the effectiveness of the proposed two-step sampling (TS). \Cref{fig:ensemble} clearly shows that RSMI with TS significantly increases the robustness for both BERT and RoBERTa. For instance, RSMI without TS shows RAcc of 64\% at $k=5$, but RSMI with TS shows RAcc of 74.1\% against TextFooler attack. We credit this robustness to the transform process of RSMI, which masks inputs and perturbs their hidden representation with Gaussian noise. This behaves as a mixture of two stochastic transformations and makes it harder for the attack algorithms to estimate the optimal perturbation direction.

\begin{table}[t]
	\centering
	\scalebox{0.85}{
	\begin{tabular}{ccccc}
	\toprule
	$\sigma$ & $M$ & $N_l$ & \bf ASR($\downarrow$) &\bf  SAcc($\uparrow$) \\ 
	\midrule
	 0.2&2& 3& 35.26& 93.15\\ 
	 0.2&2& 4& 32.73& \bf93.26\\
	 0.2&2& 5& \textbf{29.90}& 93.11\\ 
	\midrule
	 0.2 & 2 & 3 & 35.26 & \textbf{93.15}\\											
	 0.3 & 2 & 3 & 26.05 & 92.96\\ 
	 0.4 & 2 & 3 & \textbf{24.17} & 92.70\\
	 \midrule
	 0.4&2& 3 & 24.17 & \textbf{92.70}\\ 
	 0.4&3& 3 & 22.39 & 92.49\\ 
	 0.4&4& 3 & \textbf{20.99} & 92.13\\
	\bottomrule
\end{tabular}
	}
	\caption{Study on noise size ($\sigma$), number of masks ($M$), and number of noise layers ($N_l$).}
	\label{table:ablation}
\end{table}

\subsection{Impact of parameter settings}
\label{sec:ablations}

\Cref{table:ablation} shows the impact of different hyperparameters of RSMI. As observed, the overall ASR tends to decrease as we inject more noises into models by increasing noise size ($\sigma$), replacing more input words with the mask token ($M$), and adding more noise layers ($N_l$). Specifically, we observe that ASR gradually decreases as we put more noise layers in the model. Also, ASR steadily declines as we increase the standard deviation of noise. Finally, we observe that the increased number of masks effectively decreases ASR. However, we observe that these improvements in ASR come at the cost of a decreasing SAcc. 

\subsection{Training stability}
\label{sec:train_stability}

\begin{table}[t]
	\setlength{\tabcolsep}{2.0pt}
    \centering
 	\scalebox{0.88}{
    \begin{tabular}{ll|rrr}
	\toprule
        \bf Model & \bf Statistics & \bf Avg$\pm$ STD & \bf Max &\bf  Min\\ 
	\midrule
        RoBERTa-base & ~ & ~  & ~ & ~ \\ 
        +FreeLB & SAcc-$\mathcal{D}_{\text{test}}$ & $93.82 \pm 0.67$ & 94.24 & 93.05 \\ 
        ~ & RAcc & $9.21 \pm 8.57$ & 17.30 & 0.23 \\ 
		\cmidrule(r){2-5}
        +InfoBERT & SAcc-$\mathcal{D}_{\text{test}}$ & $94.14 \pm 0.10$ & 94.24 & 94.05 \\ 
        ~ & RAcc & $5.20\pm 2.24$ & 7.60 & 3.17 \\ 
		\cmidrule(r){2-5}
        +RSMI & SAcc-$\mathcal{D}_{\text{test}}$ & $92.11\pm 0.30$ & 92.40 & 91.80 \\ 
        ~ & RAcc & $71.00 \pm 2.10$ & 73.40 & 69.50 \\ 
		\midrule
        BERT-base & ~ & ~  & ~ & ~ \\ 
        +FreeLB & SAcc-$\mathcal{D}_{\text{test}}$ & $92.43 \pm 0.09$ & 92.53 & 92.36 \\ 
        ~ & RAcc & $4.09 \pm 5.57$ & 10.50 & 0.47 \\ 
		\cmidrule(r){2-5}
        +InfoBERT & SAcc-$\mathcal{D}_{\text{test}}$ & $92.68 \pm 0.23$ & 92.90 & 92.44 \\ 
        ~ & RAcc & $11.98 \pm 13.19$ & 26.40 & 0.53 \\ 
		\cmidrule(r){2-5}
        +RSMI & SAcc-$\mathcal{D}_{\text{test}}$ & $91.53 \pm 0.59$ & 92.20 & 91.07 \\ 
        ~ & RAcc & $55.69 \pm 0.84$ & 56.40 & 54.77 \\ 
		\bottomrule
\end{tabular}}
	\caption{Training stability comparison of RSMI with the baselines on IMDb. The statistics are obtained by training models three times with a different random initialization.}
	\label{table:train_stability}
\end{table}
We investigate the stability of RSMI's training dynamics. \Cref{table:train_stability} summarizes average (Avg), standard deviation (Std), max, and min values of SAcc-$\mathcal{D}_{\text{test}}$ and RAcc obtained from models trained with three different random initializations on IMDb. Note that SAcc-$\mathcal{D}_{\text{test}}$ represents SAcc for the whole test set. As shown in the table, RSMI tends to show higher stability compared to the baselines in terms of RAcc despite its stochastic nature. Despite the high and stable SAcc gains from FreeLB and InfoBERT, they tend to show significant standard deviations in RAcc and substantial gaps between the max and the min of RAcc.

\section{Conclusion}
\label{sec:conclusion}

We have proposed RSMI, a novel two-stage framework to tackle the issue of adversarial robustness of large-scale deep NLP systems. RSMI first adapts the randomized smoothing (RS) strategy for discrete text inputs and leverages a novel gradient-guided masked inference (MI) approach that reinforces the smoothing effect of RS. We have evaluated RSMI by applying it to large-scale pre-trained models on three benchmark datasets and obtain $2$ to $3$ times improvements against strong attacks in terms of robustness evaluation metrics over state-of-the-art defense methods. We have also studied the scalability of RSMI and performed extensive qualitative analyses to examine the effect of RSMI on the latent representations of the original and perturbed inputs as well as the change in its stability owing to its non-deterministic nature. Our thorough experiments and theoretical studies validate the effectiveness of RSMI as a practical approach to train adversarially robust NLP systems.

\section{Limitations}
\label{sec:limitations}
A major component of RSMI has been developed with the concept of randomized smoothing which is known to be certifiably robust within a radius of a ball around an input point. Though we have proved the robustness for the perturbed samples within this given ball, there is no theoretical guarantee that a perturbed sample will always lie within the ball. Accordingly, our study is limited to empirical validation of the effectiveness of RSMI, although it has theoretical robustness within a $L_2$ norm ball as shown in \Cref{sec:rsmi_main}. Nevertheless, certified robustness is a critical research direction for robust and reliable deployment of NLP systems to address undiscovered attacks. In our future work, we will explore the theoretical understanding of the certified-robustness of NLP systems and textual adversarial examples in-depth.

\section{Ethics Statement}
\label{sec:ethics}

The growing concern over the robustness of deep NLP systems has lead many to dedicate to develop various defense schemes but they have been typically broken by stronger attack algorithms. The proposed method demonstrates its effectiveness and potential to serve as a strong defense scheme for text classification systems. However, the disclosure of the proposed method may result in attack algorithms that are specific to target RSMI. Nonetheless, our theoretical study demonstrates that RSMI provides certifiable robustness to the NLP systems within a ball with a radius $R$, which is distinctive compared many other empirical methods, including gradient- and synonym-based works.

\section{Acknowledgements}
\label{sec:acknowledgements}
This work is partly supported by SIMTech-NTU Joint Laboratory on Complex Systems. We would like to thank the anonymous reviewers for their insightful comments. We would also like to thank Min Jeong Song for her valuable input and proofreading of the paper.

\bibliography{references}

\begin{thebibliography}{49}
\expandafter\ifx\csname natexlab\endcsname\relax\def\natexlab#1{#1}\fi

\bibitem[{Ahmed~I(1996)}]{zayed1996}
Zayed Ahmed~I. 1996.
\newblock \emph{Handbook of Function and Generalized Function Transformations}.
\newblock CRC Press, London.

\bibitem[{Alzantot et~al.(2018)Alzantot, Sharma, Elgohary, Ho, Srivastava, and
  Chang}]{alzantot-etal-2018-generating}
Moustafa Alzantot, Yash Sharma, Ahmed Elgohary, Bo-Jhang Ho, Mani Srivastava,
  and Kai-Wei Chang. 2018.
\newblock Generating natural language adversarial examples.
\newblock In \emph{Proceedings of the 2018 Conference on Empirical Methods in
  Natural Language Processing}, pages 2890--2896, Brussels, Belgium.
  Association for Computational Linguistics.

\bibitem[{Athalye et~al.(2018)Athalye, Carlini, and
  Wagner}]{athalye2018obfuscated}
Anish Athalye, Nicholas Carlini, and David Wagner. 2018.
\newblock Obfuscated gradients give a false sense of security: Circumventing
  defenses to adversarial examples.
\newblock In \emph{Proceedings of the 35th International Conference on Machine
  Learning, {ICML} 2018}.

\bibitem[{Bromiley(2003)}]{bromiley2003products}
Paul Bromiley. 2003.
\newblock Products and convolutions of gaussian probability density functions.
\newblock \emph{Tina-Vision Memo}, 3(4):1.

\bibitem[{Cohen et~al.(2019)Cohen, Rosenfeld, and Kolter}]{Cohen19_RS}
Jeremy Cohen, Elan Rosenfeld, and Zico Kolter. 2019.
\newblock \href {https://proceedings.mlr.press/v97/cohen19c.html} {Certified
  adversarial robustness via randomized smoothing}.
\newblock In \emph{Proceedings of the 36th International Conference on Machine
  Learning}, volume~97 of \emph{Proceedings of Machine Learning Research},
  pages 1310--1320. PMLR.

\bibitem[{Devlin et~al.(2019)Devlin, Chang, Lee, and Toutanova}]{bert}
Jacob Devlin, Ming-Wei Chang, Kenton Lee, and Kristina Toutanova. 2019.
\newblock {BERT}: Pre-training of deep bidirectional transformers for language
  understanding.
\newblock In \emph{Proceedings of the 2019 Conference of the North {A}merican
  Chapter of the Association for Computational Linguistics: Human Language
  Technologies, Volume 1 (Long and Short Papers)}, pages 4171--4186,
  Minneapolis, Minnesota. Association for Computational Linguistics.

\bibitem[{Dong et~al.(2021)Dong, Luu, Ji, and Liu}]{Dong2021}
Xinshuai Dong, Anh~Tuan Luu, Rongrong Ji, and Hong Liu. 2021.
\newblock Towards robustness against natural language word substitutions.
\newblock In \emph{International Conference on Learning Representations}.

\bibitem[{Däubener and Fischer(2022)}]{Sina2022}
Sina Däubener and Asja Fischer. 2022.
\newblock \href {https://doi.org/10.48550/ARXIV.2204.10839} {How sampling
  impacts the robustness of stochastic neural networks}.

\bibitem[{Ebrahimi et~al.(2018)Ebrahimi, Rao, Lowd, and
  Dou}]{Ebrahimi2018-hotflip}
Javid Ebrahimi, Anyi Rao, Daniel Lowd, and Dejing Dou. 2018.
\newblock {H}ot{F}lip: White-box adversarial examples for text classification.
\newblock In \emph{Proceedings of the 56th Annual Meeting of the Association
  for Computational Linguistics (Volume 2: Short Papers)}, pages 31--36,
  Melbourne, Australia. Association for Computational Linguistics.

\bibitem[{Fellbaum(1998)}]{wordnet}
Christiane Fellbaum. 1998.
\newblock \emph{WordNet: An Electronic Lexical Database}.
\newblock Bradford Books.

\bibitem[{Garg and Ramakrishnan(2020)}]{garg-ramakrishnan-2020-bae}
Siddhant Garg and Goutham Ramakrishnan. 2020.
\newblock {BAE}: {BERT}-based adversarial examples for text classification.
\newblock In \emph{Proceedings of the 2020 Conference on Empirical Methods in
  Natural Language Processing (EMNLP)}, pages 6174--6181, Online. Association
  for Computational Linguistics.

\bibitem[{Goyal et~al.(2022)Goyal, Doddapaneni, Khapra, and
  Ravindran}]{advNLPsurvey}
Shreya Goyal, Sumanth Doddapaneni, Mitesh~M. Khapra, and Balaraman Ravindran.
  2022.
\newblock \href {https://doi.org/10.48550/ARXIV.2203.06414} {A survey in
  adversarial defences and robustness in nlp}.

\bibitem[{Ilyas et~al.(2019)Ilyas, Santurkar, Tsipras, Engstrom, Tran, and
  Madry}]{Ilyas2019}
Andrew Ilyas, Shibani Santurkar, Dimitris Tsipras, Logan Engstrom, Brandon
  Tran, and Aleksander Madry. 2019.
\newblock \href
  {https://proceedings.neurips.cc/paper/2019/file/e2c420d928d4bf8ce0ff2ec19b371514-Paper.pdf}
  {Adversarial examples are not bugs, they are features}.
\newblock In \emph{Advances in Neural Information Processing Systems},
  volume~32. Curran Associates, Inc.

\bibitem[{Iyyer et~al.(2018)Iyyer, Wieting, Gimpel, and
  Zettlemoyer}]{iyyer-etal-2018-adversarial}
Mohit Iyyer, John Wieting, Kevin Gimpel, and Luke Zettlemoyer. 2018.
\newblock \href {https://doi.org/10.18653/v1/N18-1170} {Adversarial example
  generation with syntactically controlled paraphrase networks}.
\newblock In \emph{Proceedings of the 2018 Conference of the North {A}merican
  Chapter of the Association for Computational Linguistics: Human Language
  Technologies, Volume 1 (Long Papers)}, pages 1875--1885, New Orleans,
  Louisiana. Association for Computational Linguistics.

\bibitem[{Jin et~al.(2020)Jin, Jin, Zhou, and Szolovits}]{jin2019textfooler}
Di~Jin, Zhijing Jin, Joey~Tianyi Zhou, and Peter Szolovits. 2020.
\newblock \href {https://doi.org/10.1609/aaai.v34i05.6311} {Is bert really
  robust? a strong baseline for natural language attack on text classification
  and entailment}.
\newblock \emph{Proceedings of the AAAI Conference on Artificial Intelligence},
  34(05):8018--8025.

\bibitem[{L{\'{e}}cuyer et~al.(2019)L{\'{e}}cuyer, Atlidakis, Geambasu, Hsu,
  and Jana}]{Lecuyer19_DP}
Mathias L{\'{e}}cuyer, Vaggelis Atlidakis, Roxana Geambasu, Daniel Hsu, and
  Suman Jana. 2019.
\newblock \href {https://doi.org/10.1109/SP.2019.00044} {Certified robustness
  to adversarial examples with differential privacy}.
\newblock In \emph{2019 {IEEE} Symposium on Security and Privacy, {SP} 2019,
  San Francisco, CA, USA, May 19-23, 2019}, pages 656--672. {IEEE}.

\bibitem[{Li(2019)}]{JerryLi_RS}
Jerry Li. 2019.
\newblock \href {https://jerryzli.github.io/robust-ml-fall19/lec14.pdf} {Cse
  599-m, lecture notes of robustness in machine learning}.

\bibitem[{Li et~al.(2016)Li, Chen, Hovy, and Jurafsky}]{li2016visualizing}
Jiwei Li, Xinlei Chen, Eduard Hovy, and Dan Jurafsky. 2016.
\newblock Visualizing and understanding neural models in {NLP}.
\newblock In \emph{Proceedings of the 2016 Conference of the North {A}merican
  Chapter of the Association for Computational Linguistics: Human Language
  Technologies}. Association for Computational Linguistics.

\bibitem[{Li et~al.(2020)Li, Ma, Guo, Xue, and Qiu}]{li-etal-2020-bert-attack}
Linyang Li, Ruotian Ma, Qipeng Guo, Xiangyang Xue, and Xipeng Qiu. 2020.
\newblock {BERT}-{ATTACK}: Adversarial attack against {BERT} using {BERT}.
\newblock In \emph{Proceedings of the 2020 Conference on Empirical Methods in
  Natural Language Processing (EMNLP)}, pages 6193--6202, Online. Association
  for Computational Linguistics.

\bibitem[{Li and Qiu(2020)}]{tavat}
Linyang Li and Xipeng Qiu. 2020.
\newblock \href {https://doi.org/10.48550/ARXIV.2004.14543} {Tavat: Token-aware
  virtual adversarial training for language understanding}.

\bibitem[{Li et~al.(2021)Li, Xu, Zeng, Li, Zheng, Zhang, Chang, and
  Hsieh}]{li-etal-2021-searching}
Zongyi Li, Jianhan Xu, Jiehang Zeng, Linyang Li, Xiaoqing Zheng, Qi~Zhang,
  Kai-Wei Chang, and Cho-Jui Hsieh. 2021.
\newblock \href {https://doi.org/10.18653/v1/2021.emnlp-main.251} {Searching
  for an effective defender: Benchmarking defense against adversarial word
  substitution}.
\newblock In \emph{Proceedings of the 2021 Conference on Empirical Methods in
  Natural Language Processing}, pages 3137--3147, Online and Punta Cana,
  Dominican Republic. Association for Computational Linguistics.

\bibitem[{Liu et~al.(2018)Liu, Cheng, Zhang, and Hsieh}]{Liu18_RSE}
Xuanqing Liu, Minhao Cheng, Huan Zhang, and Cho-Jui Hsieh. 2018.
\newblock Towards robust neural networks via random self-ensemble.
\newblock In \emph{Computer Vision -- ECCV 2018}, pages 381--397, Cham.
  Springer International Publishing.

\bibitem[{Liu et~al.(2019)Liu, Ott, Goyal, Du, Joshi, Chen, Levy, Lewis,
  Zettlemoyer, and Stoyanov}]{roberta}
Yinhan Liu, Myle Ott, Naman Goyal, Jingfei Du, Mandar Joshi, Danqi Chen, Omer
  Levy, Mike Lewis, Luke Zettlemoyer, and Veselin Stoyanov. 2019.
\newblock Roberta: A robustly optimized bert pretraining approach.
\newblock \emph{arXiv preprint arXiv:1907.11692}.

\bibitem[{Loshchilov and Hutter(2019)}]{loshchilov2019decoupled}
Ilya Loshchilov and Frank Hutter. 2019.
\newblock Decoupled weight decay regularization.
\newblock In \emph{International Conference on Learning Representations}.

\bibitem[{Maas et~al.(2011)Maas, Daly, Pham, Huang, Ng, and Potts}]{imdb}
Andrew~L. Maas, Raymond~E. Daly, Peter~T. Pham, Dan Huang, Andrew~Y. Ng, and
  Christopher Potts. 2011.
\newblock Learning word vectors for sentiment analysis.
\newblock In \emph{Proceedings of the 49th Annual Meeting of the Association
  for Computational Linguistics: Human Language Technologies}, pages 142--150,
  Portland, Oregon, USA. Association for Computational Linguistics.

\bibitem[{Madry et~al.(2018)Madry, Makelov, Schmidt, Tsipras, and
  Vladu}]{madry2018towards}
Aleksander Madry, Aleksandar Makelov, Ludwig Schmidt, Dimitris Tsipras, and
  Adrian Vladu. 2018.
\newblock \href {https://openreview.net/forum?id=rJzIBfZAb} {Towards deep
  learning models resistant to adversarial attacks}.
\newblock In \emph{International Conference on Learning Representations}.

\bibitem[{Miyato et~al.(2016)Miyato, Dai, and
  Goodfellow}]{miyato2016adversarial}
Takeru Miyato, Andrew~M. Dai, and Ian Goodfellow. 2016.
\newblock \href {http://arxiv.org/abs/1605.07725} {Adversarial training methods
  for semi-supervised text classification}.
\newblock Cite arxiv:1605.07725Comment: Published as a conference paper at ICLR
  2017.

\bibitem[{Miyato et~al.(2017)Miyato, Dai, and Goodfellow}]{shen2018deep}
Takeru Miyato, Andrew~M Dai, and Ian Goodfellow. 2017.
\newblock Adversarial training methods for semi-supervised text classification.
\newblock In \emph{International Conference on Learning Representations}.

\bibitem[{Moon et~al.(2022)Moon, Joty, and Chi}]{gradmask}
Han~Cheol Moon, Shafiq Joty, and Xu~Chi. 2022.
\newblock \href {https://doi.org/10.1145/3534678.3539206} {Gradmask:
  Gradient-guided token masking for textual adversarial example detection}.
\newblock In \emph{Proceedings of the 28th ACM SIGKDD Conference on Knowledge
  Discovery and Data Mining}, KDD '22, page 3603–3613, New York, NY, USA.
  Association for Computing Machinery.

\bibitem[{Moon et~al.(2021)Moon, Mo, Lee, Lee, and Shin}]{Moon2021_Masker}
Seung~Jun Moon, Sangwoo Mo, Kimin Lee, Jaeho Lee, and Jinwoo Shin. 2021.
\newblock \href {https://ojs.aaai.org/index.php/AAAI/article/view/17601}
  {Masker: Masked keyword regularization for reliable text classification}.
\newblock \emph{Proceedings of the AAAI Conference on Artificial Intelligence},
  35(15):13578--13586.

\bibitem[{Morris et~al.(2020)Morris, Lifland, Yoo, Grigsby, Jin, and
  Qi}]{morris2020textattack}
John~X. Morris, Eli Lifland, Jin~Yong Yoo, Jake Grigsby, Di~Jin, and Yanjun Qi.
  2020.
\newblock \href {http://arxiv.org/abs/2005.05909} {Textattack: A framework for
  adversarial attacks, data augmentation, and adversarial training in nlp}.

\bibitem[{Mrk{\v{s}}i{\'c} et~al.(2016)Mrk{\v{s}}i{\'c}, {\'O}~S{\'e}aghdha,
  Thomson, Ga{\v{s}}i{\'c}, Rojas-Barahona, Su, Vandyke, Wen, and
  Young}]{mrksic-etal-2016-counter}
Nikola Mrk{\v{s}}i{\'c}, Diarmuid {\'O}~S{\'e}aghdha, Blaise Thomson, Milica
  Ga{\v{s}}i{\'c}, Lina~M. Rojas-Barahona, Pei-Hao Su, David Vandyke,
  Tsung-Hsien Wen, and Steve Young. 2016.
\newblock Counter-fitting word vectors to linguistic constraints.
\newblock In \emph{Proceedings of the 2016 Conference of the North {A}merican
  Chapter of the Association for Computational Linguistics: Human Language
  Technologies}, pages 142--148. Association for Computational Linguistics.

\bibitem[{Oppenheim et~al.(1996)Oppenheim, Willsky, and
  Nawab}]{Oppenheim1996SignalsSystems}
Alan~V. Oppenheim, Alan~S. Willsky, and S.~Hamid Nawab. 1996.
\newblock \emph{Signals \& Systems}.
\newblock Prentice-Hall, Inc., USA.

\bibitem[{Paszke et~al.(2019)Paszke, Gross, Massa, Lerer, Bradbury, Chanan,
  Killeen, Lin, Gimelshein, Antiga, Desmaison, Kopf, Yang, DeVito, Raison,
  Tejani, Chilamkurthy, Steiner, Fang, Bai, and Chintala}]{pytorch}
Adam Paszke, Sam Gross, Francisco Massa, Adam Lerer, James Bradbury, Gregory
  Chanan, Trevor Killeen, Zeming Lin, Natalia Gimelshein, Luca Antiga, Alban
  Desmaison, Andreas Kopf, Edward Yang, Zachary DeVito, Martin Raison, Alykhan
  Tejani, Sasank Chilamkurthy, Benoit Steiner, Lu~Fang, Junjie Bai, and Soumith
  Chintala. 2019.
\newblock \href
  {http://papers.neurips.cc/paper/9015-pytorch-an-imperative-style-high-performance-deep-learning-library.pdf}
  {Pytorch: An imperative style, high-performance deep learning library}.
\newblock In H.~Wallach, H.~Larochelle, A.~Beygelzimer, F.~d\textquotesingle
  Alch\'{e}-Buc, E.~Fox, and R.~Garnett, editors, \emph{Advances in Neural
  Information Processing Systems 32}, pages 8024--8035. Curran Associates, Inc.

\bibitem[{Raffel et~al.(2022)Raffel, Shazeer, Roberts, Lee, Narang, Matena,
  Zhou, Li, and Liu}]{t5_model}
Colin Raffel, Noam Shazeer, Adam Roberts, Katherine Lee, Sharan Narang, Michael
  Matena, Yanqi Zhou, Wei Li, and Peter~J. Liu. 2022.
\newblock Exploring the limits of transfer learning with a unified text-to-text
  transformer.
\newblock \emph{J. Mach. Learn. Res.}, 21(1).

\bibitem[{Ren et~al.(2019)Ren, Deng, He, and Che}]{pwws2019}
Shuhuai Ren, Yihe Deng, Kun He, and Wanxiang Che. 2019.
\newblock Generating natural language adversarial examples through probability
  weighted word saliency.
\newblock In \emph{Proceedings of the 57th Annual Meeting of the Association
  for Computational Linguistics}, pages 1085--1097, Florence, Italy.
  Association for Computational Linguistics.

\bibitem[{Salman et~al.(2019)Salman, Yang, Li, Zhang, Zhang, Razenshteyn, and
  Bubeck}]{Salman19_PRS}
Hadi Salman, Greg Yang, Jerry Li, Pengchuan Zhang, Huan Zhang, Ilya
  Razenshteyn, and S\'{e}bastien Bubeck. 2019.
\newblock Provably robust deep learning via adversarially trained smoothed
  classifiers.
\newblock In \emph{Proceedings of the 33rd International Conference on Neural
  Information Processing Systems}, Red Hook, NY, USA. Curran Associates Inc.

\bibitem[{Si et~al.(2020)Si, Zhang, Qi, Liu, Wang, Liu, and Sun}]{AdvMixUp}
Chenglei Si, Zhengyan Zhang, Fanchao Qi, Zhiyuan Liu, Yasheng Wang, Qun Liu,
  and Maosong Sun. 2020.
\newblock Better robustness by more coverage: Adversarial training with mixup
  augmentation for robust fine-tuning.
\newblock \emph{CoRR}, abs/2012.15699.

\bibitem[{Stein(1981)}]{Stein81}
Charles~M. Stein. 1981.
\newblock \href {https://doi.org/10.1214/aos/1176345632} {{Estimation of the
  Mean of a Multivariate Normal Distribution}}.
\newblock \emph{The Annals of Statistics}, 9(6):1135 -- 1151.

\bibitem[{Wang et~al.(2018)Wang, Singh, Michael, Hill, Levy, and
  Bowman}]{wang-etal-2018-glue}
Alex Wang, Amanpreet Singh, Julian Michael, Felix Hill, Omer Levy, and Samuel
  Bowman. 2018.
\newblock \href {https://doi.org/10.18653/v1/W18-5446} {{GLUE}: A multi-task
  benchmark and analysis platform for natural language understanding}.
\newblock In \emph{Proceedings of the 2018 {EMNLP} Workshop {B}lackbox{NLP}:
  Analyzing and Interpreting Neural Networks for {NLP}}, pages 353--355,
  Brussels, Belgium. Association for Computational Linguistics.

\bibitem[{Wang et~al.(2021{\natexlab{a}})Wang, Wang, Cheng, Gan, Jia, Li, and
  Liu}]{wang2021infobert}
Boxin Wang, Shuohang Wang, Yu~Cheng, Zhe Gan, Ruoxi Jia, Bo~Li, and Jingjing
  Liu. 2021{\natexlab{a}}.
\newblock Infobert: Improving robustness of language models from an information
  theoretic perspective.
\newblock In \emph{International Conference on Learning Representations}.

\bibitem[{Wang et~al.(2021{\natexlab{b}})Wang, Tang, Lou, and
  Xiong}]{Wang2021-DP}
Wenjie Wang, Pengfei Tang, Jian Lou, and Li~Xiong. 2021{\natexlab{b}}.
\newblock \href {https://doi.org/10.18653/v1/2021.naacl-main.87} {Certified
  robustness to word substitution attack with differential privacy}.
\newblock In \emph{Proceedings of the 2021 Conference of the North American
  Chapter of the Association for Computational Linguistics: Human Language
  Technologies}, pages 1102--1112, Online. Association for Computational
  Linguistics.

\bibitem[{Wang and Wang(2020)}]{Wang20-RSE}
Zhaoyang Wang and Hongtao Wang. 2020.
\newblock \href {https://doi.org/10.1007/978-3-030-55393-7_28} {Defense of
  word-level adversarial attacks via random substitution encoding}.
\newblock In \emph{Knowledge Science, Engineering and Management: 13th
  International Conference, KSEM 2020, Hangzhou, China, August 28–30, 2020,
  Proceedings, Part II}, page 312–324, Berlin, Heidelberg. Springer-Verlag.

\bibitem[{Wolf et~al.(2020)Wolf, Debut, Sanh, Chaumond, Delangue, Moi, Cistac,
  Rault, Louf, Funtowicz, Davison, Shleifer, von Platen, Ma, Jernite, Plu, Xu,
  Scao, Gugger, Drame, Lhoest, and Rush}]{wolf-etal-2020-transformers}
Thomas Wolf, Lysandre Debut, Victor Sanh, Julien Chaumond, Clement Delangue,
  Anthony Moi, Pierric Cistac, Tim Rault, Rémi Louf, Morgan Funtowicz, Joe
  Davison, Sam Shleifer, Patrick von Platen, Clara Ma, Yacine Jernite, Julien
  Plu, Canwen Xu, Teven~Le Scao, Sylvain Gugger, Mariama Drame, Quentin Lhoest,
  and Alexander~M. Rush. 2020.
\newblock \href {https://www.aclweb.org/anthology/2020.emnlp-demos.6}
  {Transformers: State-of-the-art natural language processing}.
\newblock In \emph{Proceedings of the 2020 Conference on Empirical Methods in
  Natural Language Processing: System Demonstrations}, pages 38--45, Online.
  Association for Computational Linguistics.

\bibitem[{Ye et~al.(2020)Ye, Gong, and Liu}]{safer20}
Mao Ye, Chengyue Gong, and Qiang Liu. 2020.
\newblock \href {https://doi.org/10.18653/v1/2020.acl-main.317} {{SAFER}: A
  structure-free approach for certified robustness to adversarial word
  substitutions}.
\newblock In \emph{Proceedings of the 58th Annual Meeting of the Association
  for Computational Linguistics}, pages 3465--3475, Online. Association for
  Computational Linguistics.

\bibitem[{Yoo and Qi(2021)}]{yoo-qi-2021-towards-improving}
Jin~Yong Yoo and Yanjun Qi. 2021.
\newblock \href {https://doi.org/10.18653/v1/2021.findings-emnlp.81} {Towards
  improving adversarial training of {NLP} models}.
\newblock In \emph{Findings of the Association for Computational Linguistics:
  EMNLP 2021}, pages 945--956, Punta Cana, Dominican Republic. Association for
  Computational Linguistics.

\bibitem[{Zhang et~al.(2015)Zhang, Zhao, and LeCun}]{zhang2016characterlevel}
Xiang Zhang, Junbo Zhao, and Yann LeCun. 2015.
\newblock Character-level convolutional networks for text classification.
\newblock In \emph{Advances in Neural Information Processing Systems},
  volume~28, pages 649--657. Curran Associates, Inc.

\bibitem[{Zhou et~al.(2021)Zhou, Zheng, Hsieh, Chang, and Huang}]{Zhou21-DNE}
Yi~Zhou, Xiaoqing Zheng, Cho-Jui Hsieh, Kai-Wei Chang, and Xuanjing Huang.
  2021.
\newblock \href {https://doi.org/10.18653/v1/2021.acl-long.426} {Defense
  against synonym substitution-based adversarial attacks via {D}irichlet
  neighborhood ensemble}.
\newblock In \emph{Proceedings of the 59th Annual Meeting of the Association
  for Computational Linguistics and the 11th International Joint Conference on
  Natural Language Processing (Volume 1: Long Papers)}, pages 5482--5492,
  Online. Association for Computational Linguistics.

\bibitem[{Zhu et~al.(2020)Zhu, Cheng, Gan, Sun, Goldstein, and
  Liu}]{zhu2020freelb}
Chen Zhu, Yu~Cheng, Zhe Gan, Siqi Sun, Tom Goldstein, and JJ~(Jingjing) Liu.
  2020.
\newblock \href
  {https://www.microsoft.com/en-us/research/publication/freelb-enhanced-adversarial-training-for-natural-language-understanding/}
  {Freelb: Enhanced adversarial training for natural language understanding}.
\newblock In \emph{Eighth International Conference on Learning Representations
  (ICLR)}.

\end{thebibliography}
\bibliographystyle{acl_natbib}

\appendix
\section{Stochastic stability of RSMI} 
\label{subsec:static_eval}

\begin{figure}[t]
     \centering
     \includegraphics[scale=0.60]{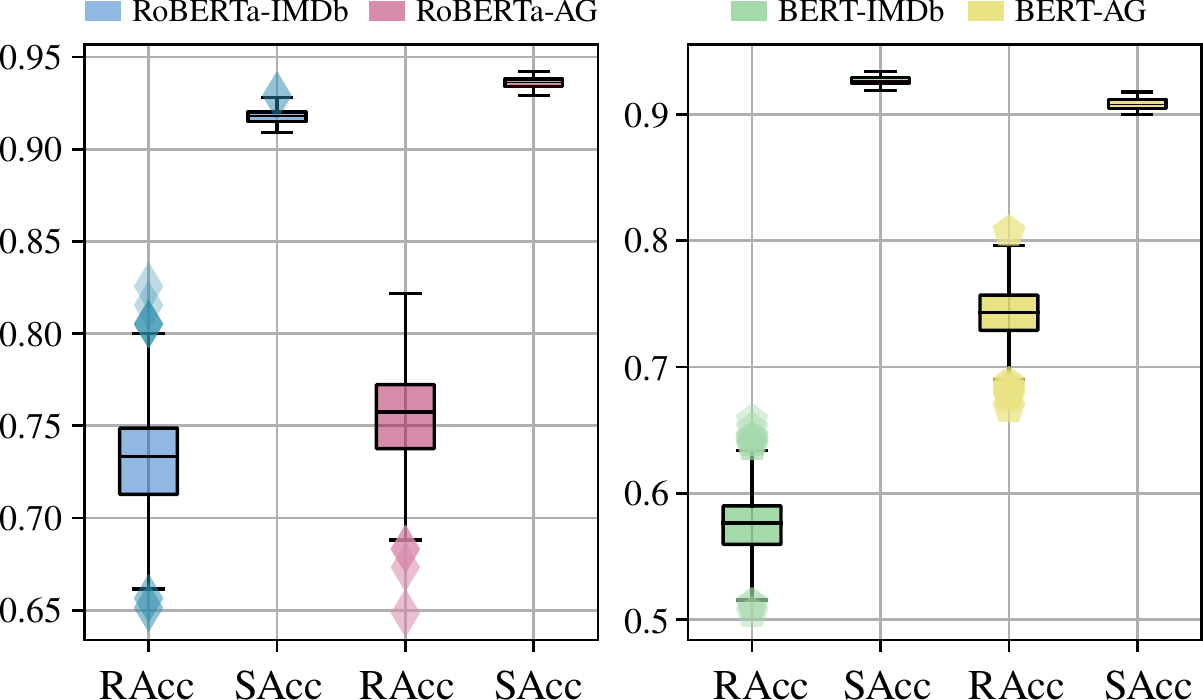}
     \caption{Stochastic stability of RSMI.}
     \label{fig:static}
\end{figure}
As RSMI is stochastic in nature, we examine its stability in classifying clean as well as adversarial samples. We first randomly draw 1,000 clean examples and evaluate them with RSMI with 1,000 independent runs. Next, we perform inference repeatedly using adversarial examples that were successful in fooling RSMI with 1,000 independent runs. As all the evaluations are independent, each evaluation involves a different noise sampling and masking process. 
\Cref{fig:static} shows that RSMI's evaluation on clean samples is significantly stable. On the other hand, most of the adversarial examples are correctly classified during each individual evaluation around the median of RSMI's RAcc. This shows that the attack success rate of adversarial attack algorithms become stochastic rather than deterministic due to RSMI's non-deterministic nature.

\section{Majority voting-based inference}
\label{sec:majority}

The inference of RSMI involves a combination of individual predictions. During evaluations in \Cref{subsec:main_exp}, RSMI is modified to draw a final decision about an input sequence by averaging logit scores of multiple Monte-Carlo samples {for a fair comparison, because the majority voting obfuscates the perturbation processes of attack algorithms by hiding model prediction scores.} However, the majority voting-based inference (\cf \Cref{alg:rsmi}) can be a practical defense method against adversarial attacks that require access to a victim model's prediction probabilities for their perturbation process since most attack algorithms require the prediction information (\eg TextFooler, PWWS, and BAE). To validate the effectiveness of the majority vote, we conduct additional experiments. As shown in \Cref{table:majority_vote}, the majority voting-based inference significantly outperforms the logit averaging approaches. 

\begin{table*}[t!]
	\setlength{\tabcolsep}{2.5pt}
	\centering
 	\scalebox{0.80}{
	\begin{tabular}{llccccccccccccc}
	\toprule
\bf Dataset &\bf  Model &\bf  SAcc ($\uparrow$) & \multicolumn{2}{c}{\bf RAcc  ($\uparrow$)} & \multicolumn{2}{c}{\bf ASR ($\downarrow$)} & \multicolumn{2}{c}{\bf AvgQ ($\uparrow$)} \\
	  &  &  & TF&PWWS & TF&PWWS & TF&PWWS \\ 
	\midrule
	\multirow{2}{*}{IMDb}
	& BERT-base    & 91.70(-0.50) & 77.20(+20.80) & 77.80(+19.10) & 15.81(-23.02) & 15.16(-21.18) & 1989(+338) & 1877(+113)\\ 
	& RoBERTa-base & 94.30(+1.3) & 81.90(+8.50) & 82.70(+6.50) & 13.15(-7.93) & 12.30(-5.77) & 2031(+114) & 1916(+53) \\ 
	\midrule
	\multirow{2}{*}{AGNews}
	& BERT-base    & 92.90(+0.20) & 85.40(+22.20) & 87.20(+11.10) & 8.07(-23.75) & 6.14(-11.77) & 572(+69) & 408(+11)\\ 
	& RoBERTa-base & 94.10(-0.20) & 86.10(+12.00) & 88.40(+6.50) & 8.50(-12.91)  & 6.06(-7.08) & 571(+41) & 407(+6)\\ 
	\midrule
	\multirow{2}{*}{QNLI}
	& BERT-base    &90.00(-0.57)& 63.60(+22.40) & 71.30(+17.10) & 29.33(-25.18) & 20.78(-19.37) & 321(+65)  & 246(+16)  \\
	& RoBERTa-base &91.89(+0.08)& 68.00(+19.00) & 76.40(+16.30) & 26.00(-20.63) & 16.86(-17.68) & 329(+63)  & 251(+11)  \\
	\bottomrule
	\end{tabular}
 	}
	\caption{Performance of the majority-voting based inference of RSMI. The round brackets next to each number denote the change of score compared to logit averaging based inference.}
	\label{table:majority_vote}
\end{table*}

\section{Natural Language Inference Task Analysis}
\label{sec:nli_task}
\Cref{table:robust_eval_nli} summarizes a performance comparison of adversarial robustness of RSMI with the baselines for NLI tasks. 

\begin{table*}[t]
	\setlength{\tabcolsep}{2.0pt}
	\centering
 	\scalebox{0.70}{
	\begin{tabular}{llc|cccc|cccc|ccccccc}
	\toprule
\bf Dataset &\bf  Model & \bf SAcc ($\uparrow$) & \multicolumn{4}{c}{\bf RAcc  ($\uparrow$)} & \multicolumn{4}{c}{\bf ASR ($\downarrow$)}&  \multicolumn{4}{c}{\bf AvgQ ($\uparrow$)}  \\
&  &  &  TF&PWWS&BAE & Avg. & TF&PWWS&BAE & Avg. & TF&PWWS&BAE  & Avg.\\ 
	\midrule
	\multirow{7}{*}{QNLI}  
	& RoBERTa-base & & & & & & & & & & & & & \\
& + Fine-Tuned         &91.90& 19.80 & 34.00 & 51.20 & 35.00 & 78.48 & 62.93 & 44.35 & 61.92 & 189  & 217  & 91    & 166  \\
& + FreeLB \free      &\bf92.10& 27.30 & 37.70 & 55.70 & 40.23 & 70.36 & 59.07 & 39.52 & 56.32 & 215  & 223  & 95    & 178  \\
& + InfoBERT \info    &91.60& 23.00 & 36.50 & 53.80 & 37.77 & 74.89 & 60.15 & 41.27 & 58.77 & 204  & 221  & 92    & 172  \\
& + SAFER \safer      &90.80& 33.80 & 45.50 & 49.70 & 43.00 & 62.82 & 49.67 & 45.26 & 52.58 & 232  & 227  & 109   & 189  \\
& + RSMI-NoMask (Our) &91.50& 34.10 & 46.80 & 50.50 & 43.80 & 62.73 & 48.73 & 45.86 & 52.44 & 218  & 225  & 113   & 185  \\
& + RSMI (Our)        &91.81&\bf 49.00 &\bf 60.10 &\bf 60.60 &\bf 56.57 &\bf 46.63 &\bf 34.54 &\bf 34.05 &\bf 38.41 &\bf 266  &\bf 240  &\bf 330   &\bf 279  \\

	\bottomrule
	\end{tabular}
 	}
	\caption{Performance comparison of adversarial robustness of RSMI with the baselines for NLI tasks. }
	\label{table:robust_eval_nli}
\end{table*}

\section{Run time analysis}
\label{sec:run_time}
\begin{table}[h]
	\setlength{\tabcolsep}{2.0pt}
	\centering
 	\scalebox{0.75}{
	\begin{tabular}{llcccccccccc}
	\toprule
	\bf Dataset &\bf  Model & \bf Train ($\downarrow$) & \bf Inference ($\downarrow$)\\
	\midrule
	\multirow{6}{*}{QNLI}
	& Fine-Tuned     & \ \,\,1.0  & \ \,\,1.0  \\ 
	& FreeLB\free   & $\times$2.8 & $\times$1.0   \\ 
	& InfoBERT\info & $\times$5.4 & $\times$1.0   \\ 
	& SAFER\safer   & $\times$1.0 & $\times$1.0 \\ 
	& RSMI NoMask (Our)    & $\times$1.9 & $\times$1.0   \\ 
	& RSMI (Our)    & $\times$1.9 & $\times$3.5   \\ 
	\bottomrule
	\end{tabular}
 	}
	\caption{Run time comparison of RSMI with the baselines.}
	\label{table:computation_time}
\end{table}

We compare the computation speed of RSMI with the baselines on the RoBERTa-base model fine-tuned on QNLI. All experiments are conducted on an Intel Xeon Gold 5218R CPU-2.10GHz processor with a single Quadro RTX 6000 GPU. For a fair comparison, the number of gradient computation steps of FreeLB and InfoBERT is set to 3 and other parameters are configured to the default settings provided by the original papers. Also, we do not include the preprocessing time of SAFER. As shown in \Cref{table:computation_time}, RSMI is approximately 1.9x slower than the Fine-Tuned model during training and 3.5x slower during inference. The latency of RSMI is mainly caused by the additional backpropagation and forward propagation for computing the gradients. The inference speed of RSMI can be improved by removing the masking step during inference, but there exist a trade-off between the inference speed and robustness as shown in \Cref{table:robust_eval}. 

\section{Experiment details}
\label{sec:parameter_setting}
\begin{table*}[t!]
	\setlength{\tabcolsep}{2.5pt}
	\begin{center}
	\begin{small}
	\scalebox{1.00}{\begin{tabular}{lllllllll}
	\toprule
	\bf Model        &   & \multicolumn{2}{c}{\bf IMDb} & \multicolumn{2}{c}{\bf AGNews}  & \multicolumn{2}{c}{\bf QNLI} \\
	        &   & RSMI & Fine-Tuned & RSMI & Fine-Tuned  & RSMI & Fine-Tuned\\
	\midrule                                  
	\multirow{14}{*}{RoBERTa} 
	&Optimizer                & AdamW   &AdamW   & AdamW   &AdamW & AdamW   &AdamW  \\
	&Batch size               & 16      &16      & 24      &24    & 36      &36     \\
	&Epochs                   & 10      &10      & 10      &10    & 10      &10   \\
	&Learning rate            & $10^-5$ &$5\times 10^-5$ & $10^-5$ &$5\times10^-5$ & $10^-5$ &$10^-5$ \\
	&Learning rate scheduler  & AL      &AL      & AL      &AL    & AL      &AL   \\
	&Maximum sequence length  & 256     &256     & 256     &256  & 256     &256    \\
	& $M$                     & 4       &- &4       &- &2       &- \\
	& $\sigma$                & 0.4     &- &0.4      &- &0.2      &- \\
	& \# Noise layers         & 3       &- &3        &- &3        &- \\
	& $\nu$                   & 1       &- &1        &- &1        &- \\
	& $k_0$                   & 5       &- &5        &- &5        &- \\
	& $k_1$                   & 50      &- &50       &- &50       &- \\
	& $\alpha$                & 0.98    &- &0.98     &- &0.98     &- \\
	& $\beta$                 & 1       &- &1        &- &1        &- \\
	\midrule
	\multirow{14}{*}{BERT} 
	&Optimizer                & AdamW    &AdamW    & AdamW    &AdamW  & AdamW    &AdamW  \\
	&Batch size               & 16      &16      & 24      &24 & 36      &36       \\
	&Epochs                   & 10      &10      & 10      &10   & 10      &10    \\
	&Learning rate            & $10^-5$ &$5\times 10^-5$ & $10^-5$ &$5\times10^-5$ & $10^-5$ &$10^-5$ \\
	&Learning rate scheduler  & AL      &AL      & AL      &AL   & AL      &AL     \\
	&Maximum sequence length  & 256     &256     & 256     &256  & 256     &256    \\
	& $M$                     & 3       &- &2       &- &2       &-  \\
	& $\sigma$                & 0.3     &- &0.2      &- &0.2      &- \\
	& \# Noise layers         & 4       &- &3        &- &3        &- \\
	& $\nu$                   & 1       &- &1        &- &1        &- \\
	& $k_0$                   & 5       &- &5        &- &5        &- \\
	& $k_1$                   & 50      &- &50       &- &50       &- \\
	& $\alpha$                & 0.98    &- &0.98     &- &0.98     &- \\
	& $\beta$                 & 1       &- &1        &- &1        &- \\
	\bottomrule
	\end{tabular}
	}
	\end{small}
	\end{center}
	\caption{Parameter settings of RSMI and the fine-tuned models. \textsc{AL} denotes the adaptive linear learning rate scheduler.}
	\label{table:parameter_settings}
\end{table*}

\subsection{Experiment environment} All of the experiments are conducted on an Intel Xeon Gold 5218R CPU-2.10GHz processor with a single Quadro RTX 6000 GPU under Python with PyTorch \citep{pytorch}.

\subsection{Models used} The models used in this work are pre-trained RoBERTa-base \citep{roberta} and BERT-base \citep{bert}, both of which have 124 million parameters. We adopt Huggingface library \citep{wolf-etal-2020-transformers} for training the models on the benchmark datasets. The huggingface code and models are all licensed under Apache 2.0, which allows for redistribution and modification. 

\subsection{Datasets used}
\Cref{table:datasets} presents the statistics of benchmarking datasets adopted in our experiments. The \textsc{IMDb} dataset contains movie reviews labeled with positive or negative sentiment labels. The \textsc{AG's News} (\textsc{AG}) dataset consists of news articles collected from more than 2,000 news sources and the samples are grouped into four coarse-grained topic classes. The objective of the NLI task is to predict the entailment relationship between a pair of sentences; whether the second sentence (\emph{Hypothesis}) is an \textit{Entailment}, a \emph{Contradiction}, or is \textit{Neutral} with respect to the first one (\emph{Premise}). The datasets are available in the public domain with custom license terms that allow non-commercial use.

\begin{table}[t]
\centering
\scalebox{1.00}{
\begin{tabular}{llllc}
\toprule
Dataset & Train & Dev & Test & \# Classes\\ 
\midrule
IMDb  & 22.5k & 2.5k & 25k  & 2 \\
AG    & 108k  & 12k & 7.6k & 4  \\
QNLI  & 105k  & 5.5k & 5.5k & 2 \\
\bottomrule
\end{tabular}
}
\caption{A summary of the benchmarking datasets.}
\label{table:datasets}
\end{table}

\subsection{Parameter settings of RSMI}
RSMI and the fine-tuned models are optimized by AdamW \adam with a linear adaptive learning rate scheduler. The maximum sequence length of input sequences is set to 256 during experiments. For the T5-Large models, we used the same parameter settings as we trained RoBERTa and BERT models except a learning rate. We set it at 0.0001 as provided in the original paper \citep{t5_model}. Further details are summarized in \Cref{table:parameter_settings}.

\subsection{Adversarial example augmentation}
\label{sec:adv_aug}

\Cref{table:adv_aug_setting} presents the number of adversarial examples used for augmenting training datasets. We generated the adversarial examples by fooling fine-tuned PLMs. To this end, we first sampled 10k clean input points from training datasets and kept the examples successfully fooling the victims. The generated adversarial examples are then augmented to each dataset. We used the same training parameters presented in \Cref{table:parameter_settings}. 

\begin{table}[t]
	\centering
	\scalebox{1.00}{
	\begin{tabular}{llr}
	\toprule
	\bf Model & \bf Dataset & \bf \# AdvEx  \\ 
	\midrule
	 BERT-base& IMDb& 9,583\\ 
	 BERT-base& AGNews& 7,463\\ 
	 RoBERTa-base& IMDb& 9,925\\ 
	 RoBERTa-base& AGNews& 7,532\\ 
	\bottomrule
\end{tabular}
	}
	\caption{Number of adversarial examples generated to augment training datasets.}
	\label{table:adv_aug_setting}
\end{table}

\subsection{Textual attack algorithm}
We employed the publicly available TextAttack library \citep{morris2020textattack} for TextFooler (TF) \citep{jin2019textfooler}, PWWS \citep{pwws2019}, and BAE \citep{garg-ramakrishnan-2020-bae} attack algorithms. We follow the default settings of each algorithm. Note that TextAttack does not include the named entity (NE) adversarial swap constraint in its PWWS implementation to extend PWWS towards a practical scenario where NE labels of input sequences are not available. As a consequence, PWWS attack in TextAttack tends to show stronger attack success rates.

\section{Proof}
\label{sec:proof_smooth}
This section provides a proof of \Cref{thrm:newradius_modified}. The sketch of the new theorem is as follows: 

Consider a simple case where a noise is added to the output of a single intermediate layer and word embeddings of a soft neural network classifier with normalization layers. The network is denoted as $F: \mathbb{R}^d \rightarrow \mathcal{P}(\mathcal{Y})$ and word embeddings of an input sequence $s$ are represented as $x$. Then, $F$ can be deemed as a composite function as follows: 
$$F = f_1 \circ f_2 = f_1(f_2(x)),$$
where $f_1$: $\mathbb{R}^{d'} \rightarrow \mathcal{P}(\mathcal{Y})$ and $f_2$: $\mathbb{R}^{d} \rightarrow \mathbb{R}^{d'}$. After injecting a noise, the new smoothed classifier can be represented as follows:
$$G = g_1 \circ g_2,$$ where $g_i$ is the \textit{Weierstrass Transform} \citep{zayed1996} of $f_i$ as stated in the following definition:
\begin{definition}
	\label{def:weier_transform}
	Denote the original soft neural network classifier as $f$, the associated smooth classifier (\textit{Weierstrass Transform} \citep{zayed1996}) can be denoted as $g$:
\begin{equation}
	\small
	g(x) = (f * \mathcal{N}(0, \sigma^2 I ))(x) = \mathop{\mathbb{E}}_{\delta \sim \mathcal{N}(0, \sigma^2 I)} [f(x+\delta)].
    \label{eq:smoothed_classifier}
\end{equation}
\end{definition}
In the following sections, we will prove $L$-Lipschitzness of $g_1$ and $g_2$. Subsequently, we will show that the output $\argmax_{y \in \gY} G(x)_y$ does not change within a certain radius of input $x$ (\cf \Cref{newradius}). Eventually, we will generalize the simplified case towards a general case where multiple layers of activations are perturbed and its radius increases exponentially as we add more noise layers to the classifier (\cf \Cref{thrm:newradius_modified}). 

We also provide justifications for the gradient-guide masking strategy (\ie MI) and show that it acts as a denoising process that enhances the smoothing effect of the proposed approach (\Cref{sec:masked_inf}). Note that we adopt \Cref{lipschitz}, \Cref{phi_lipschitz}, and \Cref{radius} from \citet{JerryLi_RS} and follow its proofs.

\subsection{Lipschitzness of the smoothed classifiers}
We will first show that $g_1$ is $\sqrt{\frac{2}{\pi \sigma^2}}$-Lipschitz in $\ell_2$ norm. Note that $g_1$ is the \textit{Weierstrass Transform} \citep{zayed1996} of a classifier $f_1$ (\cf \Cref{eq:smoothed_classifier}).
\begin{lemma}
	Let $\sigma >0$, let $h: \mathbb{R}^d \rightarrow [0, 1]^d$ be measurable, and let $H = h * \mathcal{N}(0, \sigma^2 I)$. Then H is $\sqrt{\frac{2}{\pi \sigma^2}}$-Lipschitz in $\ell_2$.
\label{lipschitz}
\end{lemma}
\begin{proof}
In $\ell_2$, we have:
\begin{equation*}
	\small
\begin{split}
&\nabla H(x) \\
& = \nabla \bigg( \frac{1}{(2\pi \sigma^2)^{\frac{d}{2}}} \int_{\mathbb{R}^d} h(t)\exp\bigg(-\frac{1}{2\sigma^2}\left\lVert x-t \right\rVert^2_2\bigg) dt\bigg) \\
& = \frac{1}{(2\pi \sigma^2)^{\frac{d}{2}}} \int_{\mathbb{R}^d} h(t) \frac{t-x}{\sigma ^2} \text{exp} (-\frac{1}{2\sigma^2}\left\lVert x-t \right\rVert^2_2) dt\,.
\end{split}
\end{equation*}
Let $v \in \mathbb{R}^d$ be a unit vector, the norm of $\nabla H(x)$ is bounded:
\begin{equation*}
	\footnotesize
\begin{split}
&| \langle v, \nabla H(x) \rangle | \\
& = \bigg| \frac{1}{(2\pi \sigma^2)^{\frac{d}{2}}} \int_{\mathbb{R}^d} h(t) \bigg\langle v, \frac{t-x}{\sigma ^2} \bigg\rangle \\
&\quad\ \exp \bigg(-\frac{1}{2\sigma^2}\left\lVert x-t \right\rVert^2_2\bigg) dt \bigg|\\
& \leq \frac{1}{(2\pi \sigma^2)^{\frac{d}{2}}} \int_{\mathbb{R}^d} \bigg| \bigg\langle v, \frac{t-x}{\sigma ^2} \bigg\rangle \bigg| \exp \bigg(-\frac{1}{2\sigma^2}\left\lVert x-t \right\rVert^2_2\bigg) dt\\
& = \frac{1}{\sigma^2} \mathop{\mathbb{E}}_{Z \sim \mathcal{N}(0, \sigma^2)}[|Z|] = \sqrt{\frac{2}{\pi\sigma^2}}\,,
\end{split}
\end{equation*}
where the second line holds since $h: \mathbb{R}^d \rightarrow [0, 1]^d$.
\end{proof} 
By \Cref{lipschitz}, $g_1$ is $\sqrt{\frac{2}{\pi \sigma^2}}$-Lipschitz.
\begin{lemma}
\label{phi_lipschitz} 
Let $\sigma > 0$, let $h: \mathbb{R}^d \rightarrow [0, 1]$, and let $H = h * \mathcal{N}(0, \sigma^2 I)$. Then the function $\Phi^{-1} (H(x))$ is $\sigma$-Lipschitz.
\end{lemma}
\begin{proof}
    Let's first consider a simple case where $\sigma=1$. Then, we have that 
	$$\nabla \Phi^{-1}(H(x)) = \frac{\nabla H(x)}{\Phi'(\Phi^{-1}(H(x)))},$$
	where $\Phi^{-1}$ is the inverse of the standard Gaussian CDF. Then, we need to show the following inequality holds for any unit vector $v$.
	\begin{align*}
		&\langle v,\nabla H(x)\rangle \leq \Phi'(\Phi^{-1}(H(x)))\\ 
		&=  \frac{1}{\sqrt{2\pi}}\exp \bigg(-\frac{1}{2}\Phi^{-1}(H(x))^2\bigg).
	\end{align*}
	By Stein's lemma \citep{Stein81}, the LHS is equal to
	$$\mathbb{E}_{X \sim \mathcal{N}(0,I)}[\langle v,X \rangle \cdot h(x+X)].$$
	We need to bound the maximum of this quantity to have the constraint that $h(x)\in [0,1]$ for all $x$ and $\mathbb{E}_{x\sim \mathcal{N}(0,I)}[h(x+X)] = p$. Let $f(z) = h(z+x)$, the problem becomes:
	\begin{equation*}
	\begin{split}
		&\max \mathop{\mathbb{E}}_{X \sim \mathcal{N}(0, I)}[\langle v, X \rangle \cdot f(X)] \\
		&\textrm{ s.t. } f(x) \in [0, 1] \quad \textrm{and}\quad  \mathop{\mathbb{E}}_{X \sim \mathcal{N}(0, I)} [f(X)]=p \,.
	\end{split}
	\end{equation*}
	The solution of the optimization problem is given by the halfspace $\ell(z) = \mathbf{1}[\langle u, z \rangle > - \Phi^{-1}(p)]$ and it is a valid solution to the problem. To show its uniqueness, let $f$ be any other possible solution and $A$ be the support of $\ell$. Then, by assumption, $\mathbb{E}_{X \sim \mathcal{N}(0, I)} [\ell(X) - f(X)] = 0$. In particular, we must have:
	\begin{align*}
	&\mathop{\mathbb{E}}_{X \sim \gN(0, I)} [(\ell(X) - f(X)) \mathbf{1}_A]\\ 
	&=  \mathop{\mathbb{E}}_{X \sim \gN(0, I)} [(\ell(X) - f(X)) \mathbf{1}_{A^C}]\,.
	\end{align*}
	However, for any $z \in A$ and $z' \in A^C$, we have that $\langle v, z \rangle \geq \langle v, z' \rangle$. Hence,
	\begin{align*}
	&\mathop{\mathbb{E}}_{X \sim \gN(0, I)} [\langle v, X \rangle(\ell(X) - f(X)) \mathbf{1}_A] \geq \\
	&\mathop{\mathbb{E}}_{X \sim \gN(0, I)} [\langle v, X \rangle(\ell(X) - f(X)) \mathbf{1}_{A^C}],
	\end{align*}
	where this uses that $\ell(z) \geq f(z)$ if $z \in A$ and $f(z) \geq \ell(z)$ otherwise. Rearranging, yields
	$$\mathop{\mathbb{E}}_{X \sim \gN(0, I)} [\langle v, X \rangle \ell(X)] \geq \mathop{\mathbb{E}}_{X \sim \gN(0, I)} [\langle v, X \rangle f(X)]$$
	as claimed. Now, we simply observe that 
	\begin{equation*}
	    \begin{split}
	        &\mathop{\mathbb{E}}_{X \sim N(0, I)} [\langle v, X \rangle \ell(X)]\\ 
			& = \mathop{\mathbb{E}}_{Z \sim \gN(0, 1)} [Z \cdot \mathbf{1}_{Z > -\Phi^{-1}(p)}]\\
	        & = \frac{1}{\sqrt{2\pi}} \int_{-\Phi^{-1}(p)}^\infty x e^{-x^2/2}dx\\
	        & = \exp \bigg(-\frac{1}{2}\Phi^{-1}(p)^2 \bigg)\\ 
			&=  \exp \bigg( -\frac{1}{2}\Phi^{-1}(H(x))^2\bigg)\,,
	    \end{split}
	\end{equation*}
	as claimed.
	
To extend the simple case to a general $\sigma$, we can take the auxiliary function $\Tilde{h}(z) = h(z/\sigma)$, and the corresponding smoothed function $\Tilde{H} = \Tilde{h} * N(0, 1)$. Then $\Tilde{H}(\sigma x) = H(x)$. By the same proof as before, $\Phi^{-1} \circ \Tilde{H}$ is 1-Lipschitz, and this immediately implies that $\Phi^{-1} \circ H$ is $\sigma$-Lipschitz.
\end{proof}

Therefore, $\Phi^{-1} (g_1(x))$ is a $\sigma$-Lipschitz smooth classifier. 

\subsection{Lipschitzness of the smoothed intermediate layers}

\begin{lemma}
\label{lemma:factor}
Let $h: \mathbb{R}^d \rightarrow \gN(0, I^{d})$ and $H = h * \gN(0, \sigma^2 I^d)$. Then, they are {$L_h$-Lipschitz and $L_H$-Lipschitz, respectively, where $L_h=\frac{1}{\sqrt{2\pi}} e^{-\frac{1}{2}}$ and $L_H = \frac{1}{1+\sigma^2} L_h$.}
\end{lemma}

\begin{proof}
{In general, a Gaussian distribution $\phi \sim \gN(\mu, \sigma^2 I)$ is $\frac{1}{\sigma^2\sqrt{2\pi}} e^{-\frac{1}{2}}$-Lipschitz, where $\phi(x)$ can be represented as follows}:
	$$\phi(x) = \frac{1}{\sqrt{2\pi\sigma^2}}\exp\bigg(-\frac{(x-\mu)^TI(x-\mu)}{2\sigma^2}\bigg).$$
	Then, the derivate of $\phi$ is given by
	\begin{small}
		\begin{align*}
			\phi'(x) = \frac{1}{\sqrt{2\pi\sigma^2}} \frac{\mu-x}{\sigma^2}\exp\bigg(-\frac{(x-\mu)^TI(x-\mu)}{2\sigma^2}\bigg)\,.
		\end{align*}
	\end{small}
	The maximum of $\lVert \phi'(x)\rVert$ can be obtained by taking the derivative of its square and set it to zero as follows: 
	\begin{small}
	\begin{align*}
    &\frac{d}{dx}\lVert \phi'(x) \rVert^2\\ 
	&= \frac{1}{2\pi\sigma^6} \bigg( -\frac{2\lVert x-\mu \rVert ^2 (x-\mu)}{\sigma^2} \exp\bigg(-\frac{\lVert x-\mu \rVert ^2}{\sigma^2}\bigg) \\
	&\quad + 2(x-\mu)\exp\bigg(-\frac{\lVert x-\mu \rVert ^2}{\sigma^2}\bigg)\bigg)=0\,.
	\end{align*}
	\end{small}
	Note that the square of the norm is a monotonic function as the norm is greater than 0. Thus, we have
	\begin{align*}
		\lVert x - \mu \rVert^2 &=  \sigma^2\,.
	\end{align*}
	This equation implies that the maximum of $\phi'$ can be found at a distance of $\sigma$ from $\mu$. For any unit vector $v \in \mathbb{R}^d$, the maximum value of $\phi'$ occurs at:
    $$x = \mu + \sigma v$$
	Subsequently, the norm of the maximum gradient is given by:
	$$\lVert\phi'(\mu + \sigma v)\rVert = \frac{1}{\sqrt{2\pi}\sigma^2} \exp\bigg(-\frac{1}{2}\bigg).$$
	Since $\lVert\phi'(x)\rVert \leq \frac{1}{\sqrt{2\pi}\sigma^2} \exp(-\frac{1}{2})$ , Lipschitz continuity of $\phi$ can be shown by the Mean Value Theorem:
	\begin{align*}
	&\lVert \phi(x) - \phi(y) \rVert \leq \sup_{x \in \mathbb{R}^d} \lVert \phi'(x) \rVert \lVert x - y \rVert \\
	&=  \frac{1}{\sqrt{2\pi}\sigma^2} \exp\bigg(-\frac{1}{2}\bigg) \lVert x - y \rVert\,.
	\end{align*}
	Therefore, $\phi \sim \gN(\mu, \sigma^2 I)$ is $\frac{1}{\sigma^2\sqrt{2\pi}} e^{-\frac{1}{2}}$-Lipschitz. Since $h$ maps to the standard normal distribution, $h(t)$ is $\frac{1}{\sqrt{2\pi}} e^{-\frac{1}{2}}$-Lipschitz. 
    
	To show the Lipschitz constraint of $H(x)$, we exploit the fact that the convolution of two Gaussian distributions $\phi_1 \sim (\mu_1, \sigma_1^2)$ and $\phi_1 \sim (\mu_2, \sigma_2^2)$ is another Gaussian distribution $\phi = \phi_1 * \phi_2 \sim (\mu_1 + \mu_2, \sigma_1^2 + \sigma_2^2)$, which could be extended to the standard multi-dimensional independent Gaussian variables with no covariance \citep{bromiley2003products}. This property leads to an equality of $H(x) = h * N(0, \sigma^2 I^d) = N(0, I^{d}) * N(0, \sigma^2 I^d) \sim N(0, (1+\sigma^2) I^d)$, which shows that $H(x)$ is $\frac{1}{\sqrt{2\pi} (1+ \sigma^2)} e^{-\frac{1}{2}I}$-Lipschitz. {Thus, $L_H = \frac{1}{1+\sigma^2} L_h$. }
\end{proof}

{\Cref{lemma:factor} implies that randomized smoothing imposes a stronger smoothness of the function, since the Lipschitz bound of the original function will be reduced by a factor of $\frac{1}{1+\sigma^2}$ as $1+\sigma^2 > 1$.} 

\subsection{Lipschitzness of the overall composite function}
\begin{lemma}
	If $f$ and $g$ are $L_1$-Lipschitz and $L_2$-Lipschitz, respectively, then the composite function $f \circ g$ is $L_1 L_2$-Lipschitz.
\label{composite}
\end{lemma}
\begin{proof}
\begin{equation*}
\begin{split}
|f\circ g (x') - f\circ g (x)| & = |f(g(x')) - f(g(x))|\\
 & \leq L_1 |g(x') - g(x)| \\
 & \leq L_1 L_2 |x' - x|\,.
\end{split}
\end{equation*}
\end{proof}
\Cref{phi_lipschitz}, \Cref{lemma:factor}, and \Cref{composite} lead to the following lemma: 
\begin{lemma}
\label{composite_lipschitz}
$\Phi^{-1}(G) = \Phi^{-1} \circ g_1 \circ g_2$ is $\Big(\frac{\sigma_1}{1+\sigma_2^2}\Big)$-Lipschitz when 
\begin{small}
$$g_1(x) = (f_1 * \mathcal{N}(0, \sigma_1^2 I ))(x) = \mathop{\mathbb{E}}_{\delta \sim \mathcal{N}(0, \sigma_1^2 I)} [f_1(x+\delta)]$$
\end{small}
and 
\begin{small}
$$g_2(x) = (f_2 * \mathcal{N}(0, \sigma_2^2 I ))(x) = \mathop{\mathbb{E}}_{\delta \sim \mathcal{N}(0, \sigma_2^2 I)} [f_2(x+\delta)]$$
\end{small}
\end{lemma}
\begin{proof}
	If a noise is only added to the input $x$, then the inverse of the standard Gaussian CDF of the smoothed classifier {, \ie\ $\Phi^{-1}\circ g(x)$,} is $\sigma_1$-Lipschitz, as stated in \Cref{phi_lipschitz}. We can consider $g$ as a composite function {, where $g = g_1'\circ g_2'$}, then $\Phi^{-1} \circ g_1'\circ g_2'(x)$ is still $\sigma_1$-Lipschitz.

	Let the Lipschitz constant of $\Phi^{-1}$ be $L_\Phi$. Note that $g_1'(x)$ is $L_1$-Lipschitz since gradients of $g_1'$ is clipped during a training phase. Also $g_2'$ is $L_2$-Lipschitz as it is a soft classifier. \Cref{composite} leads to the following observation:
$$L_\Phi L_1 L_2 = \sigma_1$$

Subsequently, we consider a case where the intermediate outputs of the composite function are perturbed. {In other words, } we also apply Weierstrass transform to the layer $f_1$ (\ie {$f_1$ to $g_1$}). Thus, $g_1$ is $\big(\frac{1}{1+\sigma_2^2} L_1\big)$-Lipschitz by \Cref{lemma:factor}.
Then, \Cref{lemma:factor} with \Cref{composite} results in the new Lipschitzness constant for $\Phi^{-1}(G)$, which is given by
$$L_\Phi \frac{1}{1+\sigma_2^2} L_1 L_2 = \frac{\sigma_1}{1+\sigma_2^2}\,.$$
\end{proof}
\subsection{Robust radius of input}

\begin{lemma}
\label{radius}
Let $m: \mathbb{R} \rightarrow \mathbb{R}$ be a monotone, invertible function. Suppose that $F:\mathbb{R}^d \to \mathcal{P}(\mathcal{Y})$ is a soft classifier, and moreover, the function $x \mapsto m(F(x)_y)$ is $L$-Lipschitz in norm $|| \cdot ||$, for every $y \in Y$. Let $a$ and $b$ are the most likely classes which are denoted as $a = \argmax_{y \in Y} G(x)_y \text{ and }~ b = \argmax_{y \in \gY\backslash{a}} G(x)_y$, respectively, and their corresponding probabilities are $p_a$ and $p_b$, then, we have that $\argmax_{y\in \gY} F(x') = a$ for all $x'$ so that $||x'-x|| < \frac{1}{2L}(m(p_a) - m(p_b))$.
\end{lemma}

\begin{proof}
As $x \mapsto m(F(x)_y)$ is $L$-Lipschitz, we know that for any $x'$ within ball $\frac{1}{2L}(m(p_a) - m(p_b))$, we have:
\begin{small}
\begin{align*}
&|m(F(x')_a) - m(F(x)_a)| = |m(F(x')_a) - m(p_a)|\\
& \leq L \lVert x' - x \rVert < \frac{1}{2}(m(p_a) - m(p_b))
\end{align*}
\end{small}
In particular, this implies that $m(F(x')_a) > \frac{1}{2}(m(p_a) + m(p_b))$. However, for any $y \neq a$, by the same logic,
\begin{align*}
&m(F(x')_y) < m(F(x)_y) + \frac{1}{2}(m(p_a) - m(p_b))\\ 
&\leq \frac{1}{2}(m(p_a) + m(p_b)) < m(F(x')_a)
\end{align*}
Hence, $\argmax_{y\in \gY} F(x') = a$.
\end{proof}

\setcounter{section}{1}
\begin{theorem}
\label{newradius}
Let $F: \mathbb{R}^d \rightarrow \mathcal{P}(\mathcal{Y})$ be any soft classifier, let $\sigma >0$, and let $G$ be its associated soft classifier, Let
$$a = \argmax_{y \in \gY} G(x)_y \quad \textrm{and}\quad  b = \argmax_{y \in \gY\backslash{a}} G(x)_y$$
be two most likely classes for $x$ according to $G$. Then, we have that $\argmax_{y \in \gY} G(x')_y = a$ for $x'$ satisfying
$$\left\lVert x' - x \right \rVert_2 \leq \frac{1+\sigma_2^2}{2\sigma_1} (\Phi^{-1} (p_a) - \Phi^{-1} (p_b))$$
\end{theorem}
\begin{proof}
	Follows from \Cref{composite_lipschitz} and \Cref{radius}.
\end{proof}
\Cref{newradius} implies that a prediction of the smoothed network is robust around the input $x$ within a radius of $\frac{1+\sigma_2^2}{2\sigma_1}(\Phi^{-1} (p_a) - \Phi^{-1} (p_b))$. {The robust radius of the proposed randomized smoothing approach is scaled} by a factor of $(1 + \sigma_2^2) > 1$. 

\subsection{Generalization to multi-layer}
\Cref{newradius} can be generalized to a multi-layer perturbation approach for a multi-layer network $F= f_1 \circ f_2 \circ \dots \circ f_L$ by repetitively applying \Cref{lemma:factor}. Then, the new smoothed classifier $G$ is $\sigma_1/\prod_{l=2}^L(1 + \sigma_l^2)$-Lipschitz and it yields \Cref{thrm:newradius_modified} with \Cref{radius} by iterative use of \Cref{lemma:factor} with \Cref{radius}.

\subsection{De-noising effect of the gradient-guided masked inference}
\label{sec:masked_inf}
The de-noising effect of the gradient-guided masked inference (MI) can be understood via its impact on the Lipschitz continuity of a soft classifier $F: \mathbb{R}^d \rightarrow \mathcal{P}(\mathcal{Y})$. Let's consider a case where we have an input sequence $s$ and its perturbed sequence $s'$. Then, their distributed representations are $x$ and $x'$, respectively. Subsequently, let $L$ be the maximum gradient for masking, then a classifier $F$ is $L$-Lipschitz, as the first derivatives is bounded by the max gradient $L$. The Lipschitz property is given by 
$$|F(x')-F(x)| \leq L|x'-x|\,.$$
The proposed gradient-guided masking process lowers the Lipschitz constant of $F$ through masking a token under a guidance of max gradient signals. It implies that $F$ will become $L'$-Lipschitz, where $L' < L$ and it can be represented as follows:
$$|F(\hat{x})-F(x)| \leq L'|\hat{x}-x| < L|x'-x|\,,$$
where $\hat{x}$ is the max gradient-guided masked word embeddings. As shown in the above equation, MI can effectively lower the upper bound of the prediction change and increases a chance of pushing the model prediction to fall into the robustness radius derived by the randomized smoothing method.

\end{document}